\newif\ifconfver
\confverfalse      %declaring conference version false
\newif\ifplainver  %declare a plain version
\plainvertrue
\ifplainver
    \confverfalse   %automatically disable conf. version argument if it's plain
\fi

\ifconfver
     \documentclass[10pt,twocolumn,twoside]{IEEEtran}
\else
    \ifplainver
        \documentclass[11pt]{article}
        \usepackage{fullpage}
    \else
        \documentclass[12pt,draftcls,onecolumn]{IEEEtran}
    \fi
\fi

\usepackage{calc,amsfonts,amssymb,amsmath,bm,url,color,theorem,graphicx,cite}
\usepackage{psfrag,subfigure,float}
\usepackage{soul}

\definecolor{orange}{RGB}{255,107,0}

\newtheorem{Fact}{Fact}
\newtheorem{Lemma}{Lemma}
\newtheorem{Prop}{Proposition}
\newtheorem{Theorem}{Theorem}
\newtheorem{Def}{Definition}
\newtheorem{Corollary}{Corollary}

\newtheorem{Assumption}{Assumption}
\theorembodyfont{\rmfamily}

\newcommand\bd{\ensuremath{{\rm bd}}}
\newcommand\inte{\ensuremath{{\rm int}}}
\newcommand{\vct}{\bm{t}}
\newcommand{\vcx}{\bm{x}}
\newcommand{\setB}{\mathcal{B}}
\newcommand{\setR}{\mathcal{R}}
\newcommand{\setW}{\mathcal{W}}
\newcommand{\Rbb}{\mathbb{R}}
\newcommand{\MVES}{\mathsf{MVES}}

\begin{document}

\bibliographystyle{IEEEtran}

%--- I do things quite strangely here to accommodate three style modes.
%--- input title and abstract here; it applies to all modes
%--- it's too complex to do authors or they are input for each mode
\newcommand{\papertitle}{
Identifiability of the Simplex Volume Minimization Criterion for Blind Hyperspectral Unmixing: The No Pure-Pixel Case
}

\newcommand{\paperabstract}{
In blind hyperspectral unmixing (HU),
the pure-pixel assumption is well-known to be powerful in enabling simple and effective blind HU solutions.
However, the pure-pixel assumption is not always satisfied in an exact sense, especially for scenarios where pixels are
%all intimately
heavily
mixed.
In the no pure-pixel case, a good blind HU approach to consider is the minimum volume enclosing simplex (MVES).
Empirical experience has suggested that MVES algorithms can perform well without pure pixels, although it was not totally clear why this is true from a theoretical viewpoint.
This paper aims to %fill the latter gap.
address the latter issue.
We develop an analysis framework wherein the perfect
endmember
identifiability of MVES is studied under the noiseless case.
We prove that MVES is indeed robust against lack of pure pixels,
as long as the pixels do not get too heavily mixed and too asymmetrically spread.
%{\blue \st{Also, our analysis reveals a surprising and counter-intuitive result, namely,
%that MVES becomes more robust against lack of pure pixels as the number of endmembers increases.}}
The theoretical results are verified by numerical simulations.
}

%--------

\ifplainver

    \date{May 30, 2014, Revised, January 9, 2015}

    \title{\papertitle}

    \author{
    $^\dag$Chia-Hsiang Lin, $^\ddag$Wing-Kin Ma, $^\dag$Wei-Chiang Li, $^\dag$Chong-Yung Chi,
    \\
    and $^\dag$ArulMurugan Ambikapathi
    \\ ~ \\
    $^\dag$Institute of Communications Engineering, National  Tsing Hua University, \\
     Taiwan, R.O.C. \\
    Emails: chiahsiang.steven.lin@gmail.com, weichiangli@gmail.com, \\  cychi@ee.nthu.edu.tw,  aareul@ieee.org
    \\ ~ \\
    $^\ddag$Department of Electronic Engineering, The Chinese University of Hong  Kong, \\
    Hong Kong \\
    Email: wkma@ieee.org
    }

    \maketitle

    \begin{abstract}
    \paperabstract
    \end{abstract}

%    \begin{center}
%        {\LARGE \papertitle} \\ ~ \\
%        Chia-Hsiang Lin, Wing-Kin Ma, Wei-Chiang Li, Chong-Yung Chi, and ArulMurugan Ambikapathi \\ ~ \\
%        Institute of Communications Engineering, National  Tsing Hua University, Taiwan, R.O.C. \\
%        Department of Electronic Engineering, The Chinese University of Hong  Kong, Hong Kong \\ ~ \\
%        \today
%    \end{center}
%
%    \section*{Abstract}
%    \paperabstract

\else
    \title{\papertitle}

    \ifconfver \else {\linespread{1.1} \rm \fi

    \author{
Chia-Hsiang Lin$^\dag$, Wing-Kin Ma$^\ddag$, Wei-Chiang
Li$^*$, Chong-Yung Chi$^\S$, and ArulMurugan
Ambikapathi$^\diamondsuit$
\thanks{Part of this paper was presented at the 38th IEEE ICASSP, Vancouver, Canada, May 26-31,
2013.
%~\cite{lin2013end}.
This work is supported by the National Science Council
(R.O.C.) under Grant NSC 102-2221-E-007-035-MY2, and partly by NTHU and
Mackay memorial hospital under Grant 100N2742E1.
%and by a General Research Fund of Hong Kong Research Grant Council (Project No. CUHK415509).
}
\thanks{$^\dag$Chia-Hsiang Lin is with Institute of Communications Engineering, National
 Tsing Hua University, Hsinchu, Taiwan 30013, R.O.C.
 E-mail: chiahsiang.steven.lin@gmail.com, Tel: +886-3-5715131X34033,
 Fax: +886-3-5751787.}
\thanks{$^\ddag$Wing-Kin Ma is with
 Department of Electronic Engineering, The Chinese University of Hong
 Kong, Shatin, N.T., Hong Kong.
 E-mail: wkma@ieee.org, Tel: +852-31634350, Fax: +852-26035558.}
\thanks{$^*$Wei-Chiang Li is with Institute of Communications Engineering, National
 Tsing Hua University, Hsinchu, Taiwan 30013, R.O.C.
 E-mail: weichiangli@gmail.com, Tel: +886-3-5715131X34033,
 Fax: +886-3-5751787.}
\thanks{$^\S$Chong-Yung Chi is the corresponding author. Address: Institute of Communications Engineering \&
 Department of Electrical Engineering, National
 Tsing Hua University, Hsinchu, Taiwan 30013, R.O.C.
 E-mail: cychi@ee.nthu.edu.tw, Tel: +886-3-5731156, Fax: +886-3-5751787.}
 \thanks{$^\diamondsuit$ArulMurugan Ambikapathi is with Institute of Communications Engineering,
National Tsing Hua University, Hsinchu, Taiwan 30013, R.O.C.
 E-mail: aareul@ieee.org, Tel: +886-3-5715131X34033,
 Fax: +886-3-5751787.}
    }

    \maketitle

    \ifconfver \else
        \begin{center} \vspace*{-2\baselineskip}
        %11th Revision, \today \\[2\baselineskip]
        \end{center}
    \fi

    \begin{abstract}
    \paperabstract
    \\\\
    \end{abstract}

    \begin{keywords}\vspace{-0.0cm}
        Hyperspectral unmixing,
        %Craig's criterion,
        minimum volume enclosing simplex,
        %endmember identifiability,
        identifiability,
        convex geometry,
        pixel purity measure
    \end{keywords}

    \ifconfver \else \IEEEpeerreviewmaketitle} \fi

 \fi

\ifconfver \else
    \ifplainver \else
        \newpage
\fi \fi
%---------------------------------------------------------------------------

\section{Introduction}

Signal, image and data processing for hyperspectral imaging
has recently received enormous attention in remote sensing~\cite{bioucas13overview,14SPM},
having numerous applications such as environmental monitoring, land mapping and classification, and object detection.
Such developments are made possible by exploiting the unique features of hyperspectral images, most notably, their high spectral resolutions.
%Signal, image and data processing for hyperspectral remote sensing
%has recently received enormous attention---with numerous applications such as environmental monitoring, land mapping and classification, and object detection---and such exciting developments are made possible by exploiting the high spectral resolution of hyperspectral images.
In this scope, blind hyperspectral unmixing (HU) is one of the topics that has aroused much interest not only from remote sensing~\cite{Jose12}, but also from other communities recently~\cite{Ken14SPM_HU,Dobigeon09,Chan2009,gillis2014fast}.
Simply speaking, the problem of blind HU is to solve a problem reminiscent of blind source separation in signal processing,
and the desired outcome is to unambiguously separate the endmember spectral signatures and their corresponding abundance maps from the observed hyperspectal scene, with no or little prior information of the mixing system.
Being given little information to solve the problem, blind HU is a challenging---but also fundamentally intriguing---problem
with many possibilities.
 %to work on with many possibilities.
%For the sake of brevity we do not review the numerous possible ways to attack blind HU;
%interested readers are referred to some recent overview articles~\cite{Jose12,Ken14SPM_HU}.
Readers are referred to some recent articles for overview of blind HU~\cite{Jose12,Ken14SPM_HU},
and here we shall not review the numerous possible ways to perform blind HU.
The focus, as well as the contribution, of this paper lie in addressing a fundamental question arising from one important blind HU approach, namely,
the minimum volume enclosing simplex (MVES) approach.

Also called simplex volume minimization or minimum volume simplex analysis (MVSA)~\cite{Li2008},
the MVES approach adopts a criterion that exploits the convex geometry structures of the observed hyperspectral data to blindly identify the endmember spectral signatures.
In the HU context
the MVES concepts were first advocated by Craig back in the 1990's~\cite{Craig1994},
%it was Craig who first advocated the MVES concepts back in the 1990's~\cite{Craig1994},
although it is interesting to note an earlier work in mathematical geology~\cite{Full81} which also described the MVES intuitions;
%see the historical note in} \cite{Ken14SPM_HU}.
see also \cite{Ken14SPM_HU} for a historical note of convex geometry, and the references therein.
In particular, Craig's work proposes the use of simplex volume as a metric for blind HU,
which is later used in some other blind HU approaches such as simplex volume maximization~\cite{Winter1999,du2008end,Chan2011} and non-negative matrix factorization~\cite{miao2007endmember}.
The MVES criterion is to minimize the volume of a simplex,
subject to constraints that the simplex encloses all hyperspectral data points.
This amounts to a nonconvex optimization problem,
and unlike the simplex volume maximization approach we do not seem to have a simple (closed-form) scheme for tackling the MVES problem.
However,
%MVES has recently gained growing attention, owing to advances in optimization.
recent advances in optimization have enabled us to handle MVES implementations efficiently.
The works in \cite{Li2008} and \cite{Chan2009} independently developed practical MVES optimization algorithms based on iterative linear approximation and alternating linear programming, respectively.
The GPU-implementation of the former is also considered very recently~\cite{agathos2014gpu}.
%Also, a very recent work considers further speed improvements by multi-GPU implementation~\cite{agathos2014gpu}.
In addition, some recent MVES algorithm designs deal with noise and outlier sensitivity issues by robust formulations,
such as the soft constraint formulation in SISAL~\cite{Dias2009} and the chance-constrained formulation in \cite{Arul2011};
the pixel elimination method in \cite{hendrix2012new} should also be noted.
We should further mention that MVES also finds application in analytical chemistry~\cite{Lopes2010}, and that fundamentally
MVES has a strong link to stochastic maximum-likelihood estimation~\cite{nascimento2012hyperspectral}.

What makes MVES special is that it seems to perform well even in the absence of pure pixels, i.e.,
pixels that are solely contributed by a single endmember.
To be more accurate, extensive simulations found that MVES may
%obtain
estimate
the ground-truth
endmembers
quite accurately in the noiseless case and without the pure-pixel assumption; see, e.g., \cite{Chan2009,nascimento2012hyperspectral,plaza2012endmember}.
At this point we should mention that while
the pure-pixel assumption is elegant and has been exploited by some other approaches, such as simplex volume maximization (also \cite{gillis2014fast} for a more recent work on near-separable non-negative matrix factorization), to arrive at remarkably simple blind HU algorithms,
it is also an arguably restrictive assumption in general.
In the HU context it has
been
suspected that MVES should be resistant to lack of pure pixels,
but it is not known to what extent MVES can guarantee perfect
endmember
identifiability under no pure pixels.
Hence, we depart from existing  MVES works, wherein improved algorithm designs are usually the theme,
and ask the following questions:
can the
endmember
identifiability of the MVES criterion in the no pure-pixel case be {\em theoretically} pinned down?
If yes, how bad (in terms of how heavy the data are mixed) can MVES withstand and where is the limit?

The contribution of this paper is theoretical.
We aim to address the aforementioned questions through analysis.
Previously, identifiability analysis for MVES was done only
%in \cite{Chan2009}, where the pure-pixel assumption is assumed.
for the pure-pixel case in \cite{Chan2009}, and for the three endmember case in the preliminary version of this paper~\cite{lin2013end}.
This paper considers the no pure-pixel case for any number of endmembers.
We prove that MVES can indeed guarantee exact
and unique
recovery of the endmembers.
The key condition for attaining such exact identifiability is that some measures concerning the pixels' purity and geometry (to be defined in Section~\ref{sec:pp_measures}) have to be above a certain limit.
The condition mentioned above is equivalent to the pure-pixel assumption for the case of two endmembers,
and is much milder than the pure-pixel assumption for the case of three endmembers or more.
%{\blue \st{Our analysis also reveals a surprising and rather counter-intuitive result---that MVES becomes more robust against lack of pure pixels as the number of endmembers increases.}}
%We will conduct numerical experiments to verify the above claim.
Numerical experiments will be conducted to verify the above claims.

This paper is organized as follows.
The problem statement is described in Section~\ref{sec:prob}.
The MVES identifiability analysis results and the associated proofs are given in Sections~\ref{sec:main_res} and \ref{sec:main_proof}, respectively.
Numerical results are provided in Section~\ref{sec:sim} to verify our theoretical claims,
and we conclude the paper in Section~\ref{sec:con}.

{\em Notations:} \
$\mathbb{R}^n$ and $\mathbb{R}^{m \times n}$ denote the sets of all real-valued $n$-dimensional vectors and $m$-by-$n$ matrices, respectively (resp.);
$\| \cdot \|$ denotes the Euclidean norm of a vector;
$\bm x^T$ denotes the transpose of $\bm x$ and the same applies to matrices;
%given a set $\mathcal{A} \subseteq \mathbb{R}^n$,
%${\rm aff}\mathcal{A}$ and ${\rm conv}\mathcal{A}$ denote the affine hull and convex hull of $\mathcal{A}$, resp. (see \cite{boyd2004convex});
%the dimension of a set $\mathcal{A} \subseteq \mathbb{R}^n$ is defined as the affine dimension of ${\rm aff}\mathcal{A}$;
%the interior and boundary of a set $\mathcal{A}$ are denoted by ${\rm int}\mathcal{A}$ and ${\rm bd}\mathcal{A}$, resp.;
given a set $\mathcal{A} \subseteq \mathbb{R}^n$,
we denote ${\rm aff}\mathcal{A}$ and ${\rm conv}\mathcal{A}$ as the affine hull and convex hull of $\mathcal{A}$, resp. (see \cite{boyd2004convex}),
${\rm int}\mathcal{A}$ and ${\rm bd}\mathcal{A}$ as the interior and boundary of $\mathcal{A}$, resp.,
and ${\rm vol}\mathcal{A}$ as the volume of $\mathcal{A}$;
the dimension of a set $\mathcal{A} \subseteq \mathbb{R}^n$ is defined as the affine dimension of ${\rm aff}\mathcal{A}$;
$\bm x \geq \bm 0$ means that $\bm x$ is elementwise non-negative;
$\bm I$ and $\bm 1$ denote an identity matrix and all-one vector of appropriate dimension, resp.;
$\bm e_i$ denotes a unit vector whose $i$th element is $[ \bm e_i ]_i = 1$ and $j$th element is $[ \bm e_i ]_j = 0$ for all $j \neq i$.

\section{Problem Statement}
\label{sec:prob}

In this section we review the background of
the MVES identifiability analysis challenge.

\subsection{Preliminaries}

Before describing the problem, some basic facts about simplex should be mentioned. A convex hull
%${\rm conv}\{ \bm b_1, \ldots, \bm b_N \} = \{ \bm x = \sum_{i=1}^N \theta_i \bm b_i ~|~ \bm \theta \geq \bm 0, \bm 1^T \bm \theta = 1 \}$,
$${\rm conv}\{ \bm b_1, \ldots, \bm b_N \} =
\left\{
\bm x = \sum_{i=1}^N \theta_i \bm b_i
~ \bigg| ~
\bm \theta \geq \bm 0, \bm 1^T \bm \theta = 1 \right\},$$
where $\bm b_1, \ldots, \bm b_N \in \mathbb{R}^{M}$, $M \geq N-1$, is called an $(N-1)$-dimensional simplex if $\bm b_1, \ldots, \bm b_N$ are affinely independent.
%Given an $(N-1)$-dimensional simplex ${\rm conv}\{ \bm b_1, \ldots, \bm b_N \} \subseteq \mathbb{R}^M$, with $M \geq N$, its volume is determined by [ref to be provided]
The volume of a simplex can be determined by~\cite{gritzmann1995largestj}
 \begin{equation} \label{eq:vol_formula}
{\rm vol}( {\rm conv}\{ \bm b_1, \ldots, \bm b_N \} ) =
\frac{1}{(N-1)!} \sqrt{ \det( \bar{\bm B}^T \bar{\bm B} ) },
\end{equation}
where $\bar{\bm B} = [~ \bm b_1 - \bm b_N, \bm b_2 - \bm b_N, \ldots, \bm b_{N-1} - \bm b_N ~] \in \mathbb{R}^{M \times (N-1)}$.
A simplex is called regular if the distances between any two vertices are the same.

\subsection{Blind HU Problem Setup}

We adopt a standard blind HU problem formulation (readers are referred to the literature, e.g., \cite{Jose12,Ken14SPM_HU}, for coverage of the underlying modeling aspects).
Concisely,
%we have an observed hyperspectral scene, whose
consider a hyperspectral scene wherein the observed
pixels can be modeled as linear mixtures of endmember spectral signatures
\begin{equation} \label{eq:basic_model}
\bm x_n = \bm A \bm s_n, \quad n=1,\ldots,L,
\end{equation}
where $\bm x_n \in \mathbb{R}^M$ denotes the $n$th pixel vector of the observed hyperspectral image, with $M$ being the number of spectral bands;
$\bm A= [~ \bm a_1, \ldots, \bm a_N ~] \in \mathbb{R}^{M \times N}$ is the endmember signature matrix, with $N$ being the number of endmembers;
$\bm s_n \in \mathbb{R}^M$ is the abundance vector of the $n$th pixel;
$L$ is the number of pixels.
The problem is to identify the unknown $\bm A$ from the observations $\bm x_1,\ldots,\bm x_L$,
thereby allowing us to unmix the abundances (also unknown) blindly.
To facilitate the subsequent problem description,
the noiseless case is assumed.
The following assumptions are standard in the blind HU context
and will be assumed throughout the paper:
(i) every abundance vector satisfies $\bm s_n \geq 0$ and $\bm 1^T \bm s_n = 1$ (i.e., the abundance non-negativity and sum-to-one constraints);
(ii) $\bm A$ has full column rank;
(iii) $[~ \bm s_1, \ldots \bm s_L ~]$ has full row rank;
(iv) $N$ is known.

\subsection{Minimum-Volume Enclosing Simplex}

\begin{figure}[htp!]
    \psfrag{a1}[Bc][Bc]{\Huge$\bm a_1$}
    \psfrag{a2}[Bc][Bc]{\Huge$\bm a_2$}
    \psfrag{a3}[Bc][Bc]{\Huge$\bm a_3$}
    \psfrag{T1}[Bc][Bc]{\huge$\mathcal{T}_1$}
    \psfrag{T2}[Bc][Bc]{\huge$\mathcal{T}_2$} \psfrag{Ta}[Bc][Bc]{\huge$\mathcal{T}_a$}
    %\psfrag{abc}[Bc][Bc]{\Huge$\mathbb{R}^M$}
    \psfrag{abc}[Bc][Bc]{\Huge }
    \centerline{\resizebox{0.5\textwidth}{!}{\includegraphics{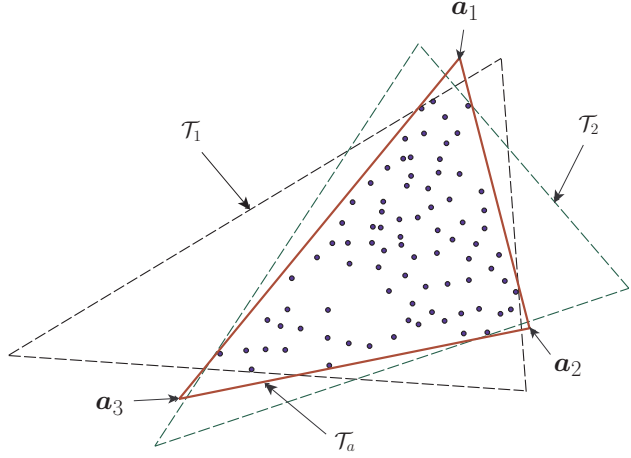}}}
    \caption{A geometrical illustration of MVES. The dots are the data points $\{ \bm x_n \}$, the number of endmembers is $N= 3$, and $\mathcal{T}_1$, $\mathcal{T}_2$ and $\mathcal{T}_a$ are data-enclosing simplices. In particular, $\mathcal{T}_a$ is actually given by $\mathcal{T}_a= {\rm conv}\{ \bm a_1, \bm a_2, \bm a_3 \}$. Visually, it can be seen that $\mathcal{T}_a$ has a smaller volume than $\mathcal{T}_1$ and $\mathcal{T}_2$.}
    \label{fig:MVES_illus}
\end{figure}

This paper concentrates on the MVES approach for blind HU.
MVES was inspired by the following intuition~\cite{Craig1994}:
if we can find a simplex that circumscribes the data points $\bm x_1,\ldots,\bm x_L$ and yields the minimum volume,
then the vertices of such a simplex should be identical to, or close to, the true endmember spectral signatures $\bm a_1, \ldots, \bm a_N$ themselves.
Figure~\ref{fig:MVES_illus} shows an illustration to support why the aforementioned intuition may be true.
Mathematically, the MVES criterion can be formulated as an optimization problem
\begin{equation} \label{eq:mves_prob}
\begin{aligned}
%\{ \hat{\bm b}_1, \ldots, \hat{\bm b}_N \} \in \arg
\min_{ \bm b_1,\ldots, \bm b_N \in \mathbb{R}^M } & ~ {\rm vol}({\rm conv}\{\bm b_1,\ldots, \bm b_N\}) \\
{\rm s.t.} & ~ \bm x_n \in {\rm conv}\{\bm b_1,\ldots, \bm b_N\}, ~n=1,\ldots,L,
\end{aligned}
\end{equation}
wherein the solution of problem~\eqref{eq:mves_prob} is used as an estimate of $\bm A$.
%Algorithms for handling problem~\eqref{eq:mves_prob} have been previously proposed;
% %and readers are referred to the literature \cite{Chan2009,Li2008,Lopes2010,nascimento2012hyperspectral};
Problem~\eqref{eq:mves_prob} is NP-hard in general~\cite{packer2002np}; this means that the optimal MVES solution is unlikely to be computationally tractable for any arbitrarily given $\{ \bm x_n \}_{n=1}^L$.
Notwithstanding, %there is growing empirical evidence that suggests
it was found
that
carefully designed algorithms for handling problem~\eqref{eq:mves_prob}, though being generally suboptimal in view of the NP-hardness of problem~\eqref{eq:mves_prob},
can practically yield satisfactory endmember identification performance;
see, e.g., \cite{Chan2009,Li2008,Lopes2010,nascimento2012hyperspectral}, and
also \cite{miao2007endmember,Arul2011,Dias2009,hendrix2012new} for the noisy case.
%{\blue \st{Our interest is not with MVES algorithm design.}}
In this paper, we do not consider MVES algorithm design.
Instead, we
%consider
study
%{\blue This motivates us to consider}
the following fundamental, and very important, question: {\em When will
%the MVES criterion in \eqref{eq:mves_prob}
the MVES problem~\eqref{eq:mves_prob}
provide %a
an optimal
solution that is exactly and uniquely given by the true endmember matrix $\bm A$ (up to a permutation)?}

%\begin{figure}[htp!]
%    \psfrag{a1}[Bc][Bc]{\Huge$\bm a_1$}
%    \psfrag{a2}[Bc][Bc]{\Huge$\bm a_2$}
%    \psfrag{a3}[Bc][Bc]{\Huge$\bm a_3$}
%    \psfrag{T1}[Bc][Bc]{\huge$\mathcal{T}_1$}
%    \psfrag{T2}[Bc][Bc]{\huge$\mathcal{T}_2$} \psfrag{Ta}[Bc][Bc]{\huge$\mathcal{T}_a$}
%    %\psfrag{abc}[Bc][Bc]{\Huge$\mathbb{R}^M$}
%    \psfrag{abc}[Bc][Bc]{\Huge }
%    \centerline{\resizebox{0.5\textwidth}{!}{\includegraphics{fig/NoPurePixelIsOK.eps}}}
%    \caption{A geometrical illustration of MVES. The dots are the data points $\{ \bm x_n \}$, the number of endmembers is $N= 3$, and $\mathcal{T}_1$, $\mathcal{T}_2$ and $\mathcal{T}_a$ are data-enclosing simplices. In particular, $\mathcal{T}_a$ is actually given by $\mathcal{T}_a= {\rm conv}\{ \bm a_1, \bm a_2, \bm a_3 \}$. Visually, it can be seen that $\mathcal{T}_a$ has a smaller volume than $\mathcal{T}_1$ and $\mathcal{T}_2$.}
%    \label{fig:MVES_illus}
%\end{figure}

%\subsection{Theoretical Identifiability of MVES}

It is known that MVES uniquely identifies $\bm A$ if the pure-pixel assumption holds~\cite{Chan2009},
that is, if, for each $i \in \{ 1,\ldots, N \}$, there exists an abundance vector $\bm s_n$ such that $\bm s_n = \bm e_i$.
However, empirical evidence has suggested that even when the pure-pixel assumption does not hold, MVES
{(more precisely, approximate MVES by the existing algorithms)}
may still be able to uniquely identify $\bm A$.
%{\blue \st{At present, there are few works %that proves whether this is also true in theory, and to what extent the pure pixel assumption can be violated.
%on providing provable MVES identifiability results in the no pure-pixel case,}}
%{\blue To the best of our knowledge, there is no existing work that tackles
%provable MVES identifiability in the no pure-pixel case and for general $N$,}
%and this paper aims at doing so.
%{\blue \st{At present, there are few works %that proves whether this is also true in theory, and to what extent the pure pixel assumption can be violated.
%on providing provable MVES identifiability results in the no pure-pixel case, and this paper aims at doing so.}}
In this paper, we aim at analyzing the endmember identifiability of MVES in the no pure-pixel case.

\section{Main Results}
\label{sec:main_res}

This section describes the main results of our MVES identifiability analysis.
As will be seen soon, MVES identifiability in the no pure-pixel case depends much on the level of  ``pixel purity'' of the observed data set.
To this end, we need to precisely quantify what ``pixel purity'' is.
The first subsection will introduce two pixel purity measures.
The second subsection will then present the main results,
%and discuss their implications in practice.
and the third subsection will discuss their practical implications.

\subsection{Pixel Purity Measures}
\label{sec:pp_measures}

%As will be shown in the next subsection, the MVES identifiability in the no pure-pixel case depends much on the level of  ``pixel purity'' of the observed data  set.
%To this end, we need to precisely quantify what ``pixel purity'' is.
%One measure is given as follows

A natural way to quantify pixel purity is to use the following measure
\begin{equation} \label{eq:rho_def}
%\rho = \sup\{ \| \bm s_n \| ~|~ n=1,\ldots, L \}.
{ \rho = \max_{n=1,\ldots,L} \| \bm s_n \|.}
\end{equation}
Eq.~\eqref{eq:rho_def} will be called the {\em best pixel purity level} in the sequel.
A large $\rho$ implies that there exist abundance vectors whose purity is high, while a small $\rho$ indicates more heavily mixed data.
To see it, observe that $\| \bm s \| \leq 1$ for any  $\bm s \geq \bm 0$, $\bm 1^T \bm s = 1$, and equality holds if and only if $\bm s = \bm e_k$ for any $k$; that is, a pure pixel.
Moreover, it can be shown that $\frac{1}{\sqrt{N}} \leq \| \bm s \|$ for any  $\bm s \geq \bm 0$, $\bm 1^T \bm s = 1$, and equality holds if and only if $\bm s = \frac{1}{N} \bm 1$; that is, a heavily mixed pixel.
Without loss of generality {(w.l.o.g.)}, we may assume
\[ \frac{1}{\sqrt{N}} <  \rho \leq 1, \]
where we rule out $\rho= \frac{1}{\sqrt{N}}$, which implies $\bm s_1 = \ldots = \bm s_L = \frac{1}{N} \bm 1$ and leads to a pathological case.

The previously defined pixel purity level reflects the best abundance purity among all the pixels, but says little on how the pixels are spread geometrically with respect to (w.r.t.) the various endmembers.
We will also require another measure, defined as follows
\begin{equation} \label{eq:gamma_def}
\gamma = \sup\{ { r \leq 1} ~|~ \mathcal{R}(r) \subseteq {\rm conv}\{ \bm s_1, \ldots, \bm s_L \} \},
\end{equation}
where
%\begin{equation} \label{eq:Rr_def}
%\mathcal{R}(r) = \{ \bm s \in \mathbb{R}^N ~|~ \| \bm s \| \leq r \} \cap {\rm conv} \{ \bm e_1, \ldots, \bm e_N \}.
%\end{equation}
\begin{align}
 \mathcal{R}(r)
& =  { \{ \bm s \in {\rm conv} \{ \bm e_1, \ldots, \bm e_N \} ~|~ \| \bm s \| \leq r \} } \nonumber \\
& = \{ \bm s \in \mathbb{R}^N ~|~ \| \bm s \| \leq r \} \cap {\rm conv} \{ \bm e_1, \ldots, \bm e_N \}.
\label{eq:Rr_def}
\end{align}
We call \eqref{eq:gamma_def} the {\it uniform pixel purity level}; the reason for this will be illustrated soon.
%(provide an illustration to give some physical feel to readers).
It can be shown that $$\frac{1}{\sqrt{N}} \leq \gamma \leq \rho.$$
Also, if $\gamma= 1$, then the pure-pixel assumption is shown to hold.

\begin{figure}[htp!]
    \psfrag{e1}[Bc][Bc]{\Huge$\bm e_1$}
    \psfrag{e2}[Bc][Bc]{\Huge$\bm e_3$}
    \psfrag{e3}[Bc][Bc]{\Huge$\bm e_2$}
    \psfrag{R1}[Bc][Bc]{\huge${\mathcal R}(r)$}
    \psfrag{R11}[Bc][Bc]{\huge$(r \leq 1/{\sqrt 2})$}
    \psfrag{R2}[Bc][Bc]{\huge${\mathcal R}(r)$}
    \psfrag{R22}[Bc][Bc]{\huge$(r > 1/{\sqrt 2})$}
    %\psfrag{abc}[Bc][Bc]{\Huge$\mathbb{R}^3$}
    \psfrag{abc}[Bc][Bc]{}
    %\centerline{\resizebox{0.7\textwidth}{!}{\includegraphics{fig/FigC.eps}}}
    \begin{center}
        \subfigure[]{\resizebox{0.45\textwidth}{!}{\includegraphics{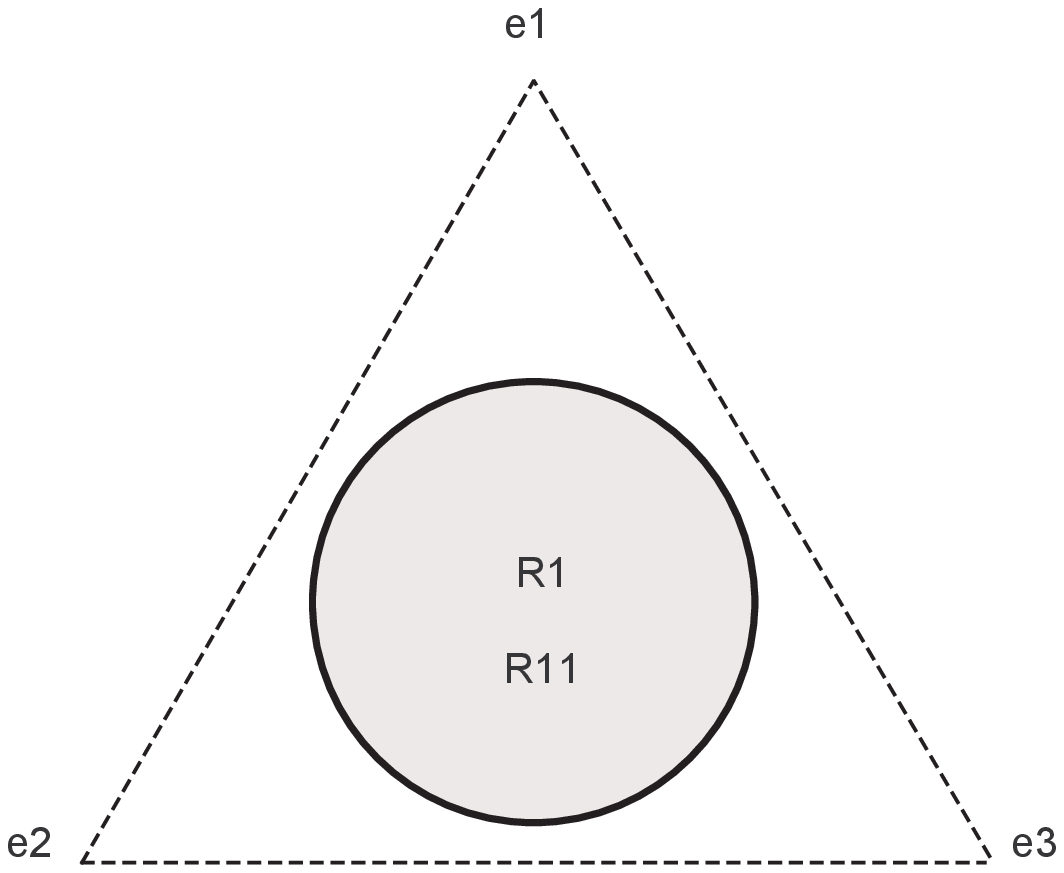}}}
        \subfigure[]{\resizebox{0.45\textwidth}{!}{\includegraphics{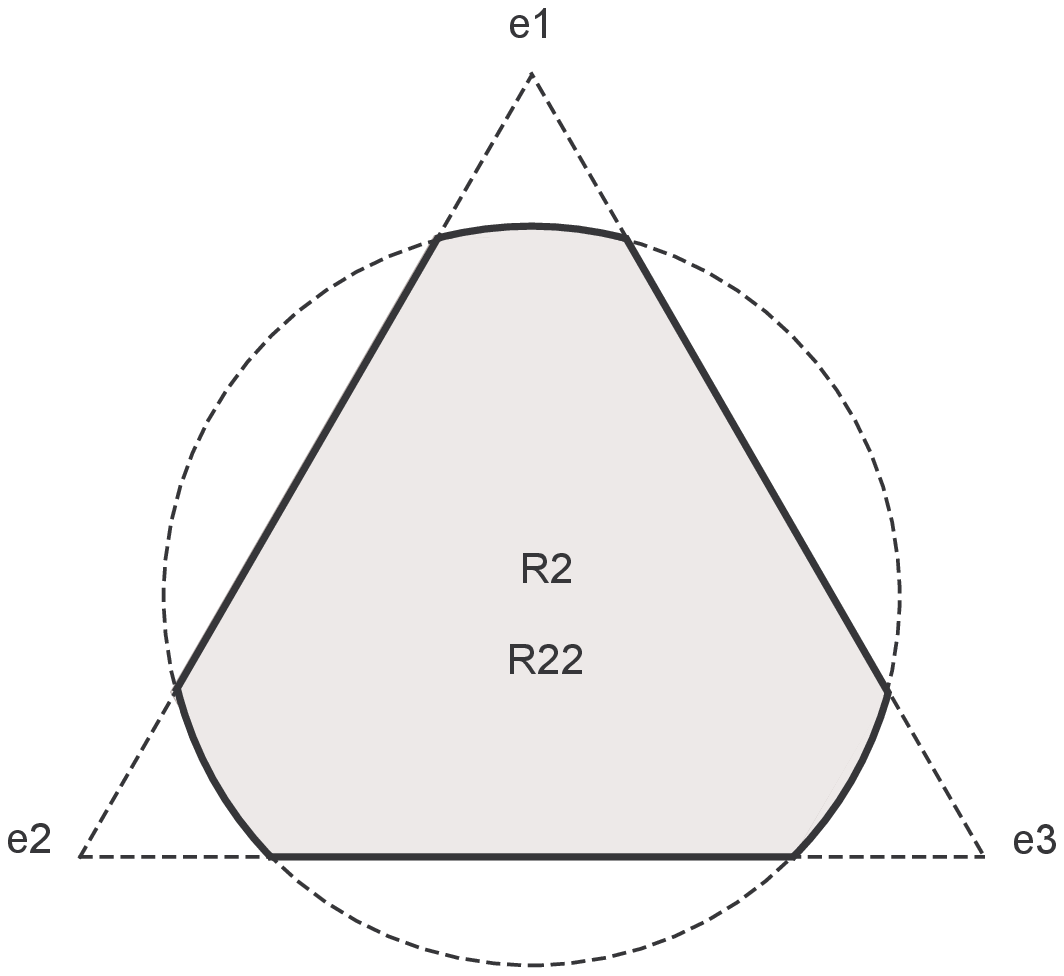}}}
    \end{center}
    \caption{A geometrical illustration of $\mathcal{R}(r)$ in \eqref{eq:Rr_def} for $N= 3$. We view $\mathcal{R}(r)$ by adjusting the viewpoint to be perpendicular to the affine hull of $\{ \bm e_1,\bm e_2,\bm e_3 \}$.}
    \label{fig:Rr_illus}
\end{figure}

To understand the differences between the pixel purity measures in \eqref{eq:rho_def} and \eqref{eq:gamma_def}, we first illustrate how $\mathcal{R}(r)$ looks like in Figure~\ref{fig:Rr_illus}.
As can be seen (and as will be shown), $\mathcal{R}(r)$ is a ball on the affine hull ${\rm aff}\{ \bm e_1,\ldots, \bm e_N \}$ if $r \leq 1/\sqrt{N-1}$.
Otherwise, $\mathcal{R}(r)$ takes a shape like a vertices-cropped version of the unit simplex ${\rm conv}\{ \bm e_1,\ldots, \bm e_N \}$.
In addition, it can be shown that \eqref{eq:rho_def} equals
\begin{equation*}
\rho = \inf\{ r ~|~ {\rm conv}\{ \bm s_1, \ldots, \bm s_L \} \subseteq \mathcal{R}(r)  \}.
\end{equation*}
%note that the above expression of $\rho$ is  \eqref{eq:gamma_def}.
In Figure~\ref{fig:abundances}, we give several examples with the abundances.
From the figures, an interesting observation is that $\mathcal{R}(\rho)$ serves as a smallest $\mathcal{R}(r)$ that circumscribes the abundance convex hull ${\rm conv}\{ \bm s_1,\ldots, \bm s_L \}$,
while $\mathcal{R}(\gamma)$ serves as a largest $\mathcal{R}(r)$ that is inscribed in ${\rm conv}\{ \bm s_1,\ldots, \bm s_L \}$.
Moreover, we see that if the abundances are spread in a relatively symmetric manner w.r.t. all the endmembers, then $\rho$ and $\gamma$ are similar;
this is the case with Figures~\ref{fig:abundances}(a)-\ref{fig:abundances}(c).
However, $\rho$ and $\gamma$ can be quite different if the abundances are asymmetrically spread;
this is the case with Figure~\ref{fig:abundances}(d) where some endmembers have pixels of high purity but some do not.
Hence, the uniform pixel purity level $\gamma$ quantifies a pixel purity level that applies uniformly to {\em all} the endmembers, not just to the best.

\ifconfver
    \begin{figure*}
\else
    \begin{figure}[htp!]
\fi
\begin{center}
\subfigure[$\gamma < 1/\sqrt{2}, \rho < 1/\sqrt{2}$]{
\psfrag{e1}[Bc][Bc]{\small$\bm e_1$}
\psfrag{e2}[Bc][Bc]{\small$\bm e_2$}
\psfrag{e3}[Bc][Bc]{\small$\bm e_3$}
%\psfrag{Te}[Bc][Bc]{\huge$\Te$}
\psfrag{ALL}[Bc][Bc]{\small$\mathcal{R}(\gamma)= {\rm conv}\{\bm s_1,\ldots,\bm s_L\}$}
\psfrag{ALL2}[Bc][Bc]{\small$=\mathcal{R}(\rho)$}
\psfrag{Rrho}[Bc][Bc]{\small$\mathcal{R}(\rho)$}
\psfrag{Rr}[Bc][Bc]{\small$\mathcal{R}(\gamma)$}
\psfrag{convSL}[Bc][Bc]{\small${\rm conv}\{\bm s_1,\ldots,\bm s_L\}$}
   \includegraphics[scale =0.6] {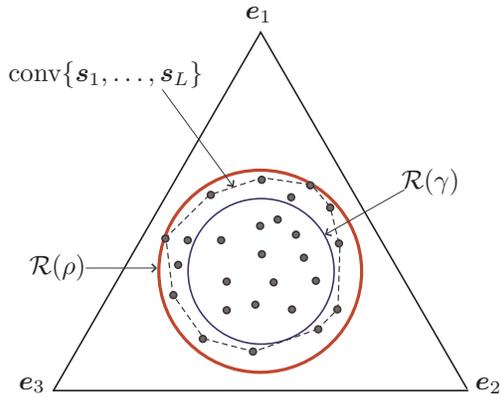}
 }\label{subFIG_3a}
\subfigure[$\gamma > 1/\sqrt{2}, \rho > 1/\sqrt{2}$]{
\psfrag{e1}[Bc][Bc]{\small$\bm e_1$}
\psfrag{e2}[Bc][Bc]{\small$\bm e_2$}
\psfrag{e3}[Bc][Bc]{\small$\bm e_3$}
%\psfrag{Te}[Bc][Bc]{\huge$\Te$}
\psfrag{ALL}[Bc][Bc]{\small$\mathcal{R}(\gamma)= {\rm conv}\{\bm s_1,\ldots,\bm s_L\}$}
\psfrag{ALL2}[Bc][Bc]{\small$=\mathcal{R}(\rho)$}
\psfrag{Rrho}[Bc][Bc]{\small$\mathcal{R}(\rho)$}
\psfrag{Rr}[Bc][Bc]{\small$\mathcal{R}(\gamma)$}
\psfrag{convSL}[Bc][Bc]{\small${\rm conv}\{\bm s_1,\ldots,\bm s_L\}$}
   \includegraphics[scale =0.6] {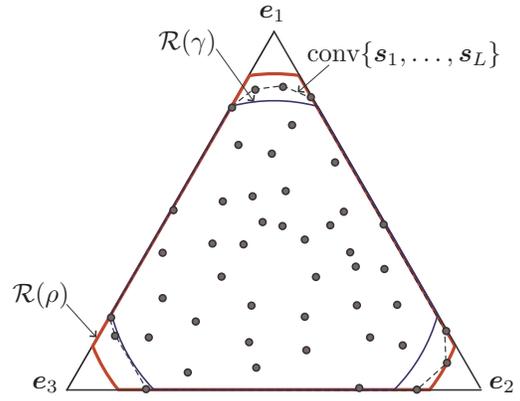}
 }\label{subFIG_3b}

\subfigure[$\gamma=\rho=1$]{
\psfrag{e1}[Bc][Bc]{\small$\bm e_1$}
\psfrag{e2}[Bc][Bc]{\small$\bm e_2$}
\psfrag{e3}[Bc][Bc]{\small$\bm e_3$}
%\psfrag{Te}[Bc][Bc]{\huge$\Te$}
\psfrag{ALL}[Bc][Bc]{\small${\rm conv}\{\bm s_1,\ldots,\bm s_L\}$}
\psfrag{ALL2}[Bc][Bc]{\small$=\mathcal{R}(\rho)=\mathcal{R}(\gamma)$}
\psfrag{Rrho}[Bc][Bc]{\small$\mathcal{R}(\rho)$}
\psfrag{Rr}[Bc][Bc]{\small$\mathcal{R}(\gamma)$}
\psfrag{convSL}[Bc][Bc]{\small${\rm conv}\{\bm s_1,\ldots,\bm s_L\}$}
   \includegraphics[scale =0.6] {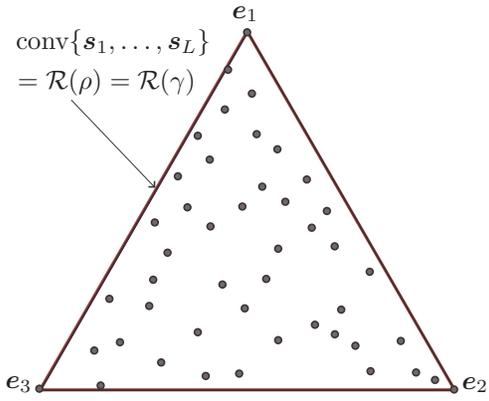}
 }\label{subFIG_3c}
%\subfigure[$\gamma=\rho=1$]{
%\psfrag{e1}[Bc][Bc]{\small$\bm e_1$}
%\psfrag{e2}[Bc][Bc]{\small$\bm e_2$}
%\psfrag{e3}[Bc][Bc]{\small$\bm e_3$}
%%\psfrag{Te}[Bc][Bc]{\huge$\Te$}
%\psfrag{ALL}[Bc][Bc]{\small$\mathcal{R}(\gamma)= {\rm conv}\{\bm s_1,\ldots,\bm s_L\}$}
%\psfrag{ALL2}[Bc][Bc]{\small$=\mathcal{R}(\rho)$}
%\psfrag{Rrho}[Bc][Bc]{\small$\mathcal{R}(\rho)$}
%\psfrag{Rr}[Bc][Bc]{\small$\mathcal{R}(\gamma)$}
%\psfrag{convSL}[Bc][Bc]{\small${\rm conv}\{\bm s_1,\ldots,\bm s_L\}$}
%   \includegraphics[scale =0.6] {fig/fig3c.eps}
% }\label{subFIG_3c}
\subfigure[$\gamma<1/\sqrt{2}$, $\rho > 1/\sqrt{2}$]{
\psfrag{e1}[Bc][Bc]{\small$\bm e_1$}
\psfrag{e2}[Bc][Bc]{\small$\bm e_2$}
\psfrag{e3}[Bc][Bc]{\small$\bm e_3$}
%\psfrag{Te}[Bc][Bc]{\huge$\Te$}
\psfrag{ALL}[Bc][Bc]{\small$\mathcal{R}(\gamma)= {\rm conv}\{\bm s_1,\ldots,\bm s_L\}$}
\psfrag{ALL2}[Bc][Bc]{\small$=\mathcal{R}(\rho)$}
\psfrag{Rrho}[Bc][Bc]{\small$\mathcal{R}(\rho)$}
\psfrag{Rr}[Bc][Bc]{\small$\mathcal{R}(\gamma)$}
\psfrag{convSL}[Bc][Bc]{\small${\rm conv}\{\bm s_1,\ldots,\bm s_L\}$}
   \includegraphics[scale =0.6] {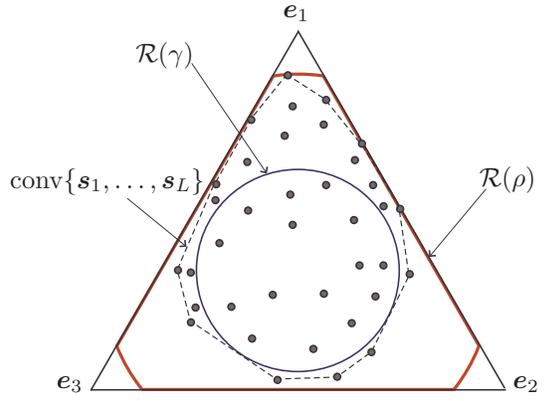}
 }\label{subFIG_3d}
\end{center}
\caption{Examples with the abundance distributions and the corresponding best and uniform pixel purity levels.}
\label{fig:abundances}
%\end{figure}
\ifconfver
    \end{figure*}
\else
    \end{figure}
\fi

\subsection{Provable MVES Identifiability}

%This section describes the main analysis results and their implications.
Our provable MVES identifiability results are described as follows.
To facilitate our analysis, consider the following definition.
\begin{Def}[minimum volume enclosing simplex] \label{def:mves}
Given an $m$-dimensional set $\mathcal{U} \subseteq \mathbb{R}^n$,
the notation ${\sf MVES}(\mathcal{U})$ denotes the set that collects all $m$-dimensional minimum volume simplices that enclose $\mathcal{U}$ and lie in ${\rm aff}\mathcal{U}$.
\end{Def}
Now, let \begin{align*} \mathcal{T}_e & = {\rm conv}\{ \bm e_1, \ldots, \bm e_N \} \subseteq \mathbb{R}^N,  \\
\mathcal{T}_a & = {\rm conv}\{ \bm a_1, \ldots, \bm a_N \} \subseteq \mathbb{R}^M,
\end{align*}
denote the $(N-1)$-dimensional unit simplex and the {endmembers'} simplex, respectively.
Also, for convenience, let
\[ \mathcal{X}_L = \{ \bm x_1, \ldots \bm x_L \}, \quad \mathcal{S}_L = \{ \bm s_1, \ldots \bm s_L \}, \]
{denote the sets of all the observed hyperspectral pixels and abundance vectors, resp.,
and %recall
note their dependence
%the linear model
$\bm x_n = \bm A \bm s_n$ as described in \eqref{eq:basic_model}.}
Under the above definition,
the exact and unique identifiability problem of the MVES criterion in \eqref{eq:mves_prob} can be posed as a problem of finding conditions under which
\[
{\sf MVES}(\mathcal{X}_L) = \{ \mathcal{T}_a \}.
\]
%where $\mathcal{X}_L = \{ \bm x_1, \ldots \bm x_L \}$.
%Let $\mathcal{S}_L = \{ \bm s_1, \ldots \bm s_L \}$.

Our first result reveals that the MVES perfect identifiability does not depend on $\bm A$ (as far as $\bm A$ has full column rank):
\begin{Prop} \label{prop:mut}
${\sf MVES}( \mathcal{X}_L ) = \{ \mathcal{T}_a \}$ if and only if ${\sf MVES}( \mathcal{S}_L ) = \{ \mathcal{T}_e \}$.
\end{Prop}
The proof of Proposition~\ref{prop:mut}, as well as those of the theorems to be presented, will be provided in the next section.
Proposition~\ref{prop:mut} suggests that to analyze the perfect MVES identifiability w.r.t. the observed pixel vectors, it is equivalent to analyze the perfect MVES identifiability w.r.t. the abundance vectors.
One may expect that perfect identifiability cannot be achieved for too heavily mixed pixels.
We prove that this is indeed true.
\begin{Theorem} \label{thm:nec}
Assume $N \geq 3$. If ${\sf MVES}(\mathcal{S}_L) = \{ \mathcal{T}_e \}$, then the best pixel purity level must satisfy $\rho > \frac{1}{\sqrt{N-1}}$.
\end{Theorem}
%Again, Theorem~\ref{thm:nec} will be shown later.
To get some idea, consider the example in Figure~\ref{fig:abundances}(a). Since Figure~\ref{fig:abundances}(a) does not satisfy the condition in Theorem~\ref{thm:nec},
%it stands no chance in exact recovery of the true endmembers.
it fails to provide exact recovery of the true endmembers.
Theorem~\ref{thm:nec} is only a necessary perfect identifiability condition.
We also prove a sufficient perfect identifiability condition, described as follows:
\begin{Theorem} \label{thm:suff}
Assume $N \geq 3$. If the uniform pixel purity level satisfies $\gamma > \frac{1}{\sqrt{N-1}}$, then ${\sf MVES}(\mathcal{S}_L) = \{ \mathcal{T}_e \}$.
\end{Theorem}
%Theorem~\ref{thm:suff} will also be shown later.
Among the four examples in Figure~\ref{fig:abundances},
Figure~\ref{fig:abundances}(b) and Figure~\ref{fig:abundances}(c) are cases that satisfy the condition in Theorem~\ref{thm:suff} and achieve exact and unique recovery of the true endmembers.

It is worthwhile to emphasize that
the sufficient identifiability condition in Theorem~\ref{thm:suff} is much milder than the pure-pixel assumption (which is equivalent to $\gamma=1$) for $N \geq 3$.
%{\blue\st{, especially when $N$ is large}}.
%{\blue \st{In fact, for a very large $N$, the pixel purity requirement $1/\sqrt{N-1}$ vanishes to zero}
{In fact, the pixel purity requirement $1/\sqrt{N-1}$ diminishes as $N$ increases}---which
%implies
{seems to suggest}
that MVES can handle
%increasingly
{more heavily}
mixed cases as the number of endmembers increases.
%This is rather a counter-intuitive conclusion.
%{\blue \st{While such a conclusion is rather counter-intuitive,
%it for the first time provides a theoretical justification on why
%%MVES can be robust against lack of pure pixels in practice.
%previous works found that MVES or related algorithms can be so robust against lack of pure pixels in practice.}}
{Thus, Theorem~\ref{thm:suff} provides a theoretical justification on the robustness of MVES against lack of pure pixels.}

{
One may be curious
%about the idea behind proving Theorem~\ref{thm:suff}.
about how Theorem~\ref{thm:suff} is proven.
Essentially, the idea lies in finding a connection between the MVES identifiability conditions of $\mathcal{S}_L$ and $\mathcal{R}(\gamma)$ [cf. \eqref{eq:gamma_def}-\eqref{eq:Rr_def}].
In particular, it is shown that if ${\sf MVES}(\mathcal{R}(\gamma)) = \{ \mathcal{T}_e \}$, then ${\sf MVES}(\mathcal{S}_L) = \{ \mathcal{T}_e \}$.
Subsequently, the problem is to pin down the MVES identifiability condition of $\mathcal{R}(r)$.
%As the key, and arguably most challenging, part of the proof, we prove the following result:
This turns out to be the core part of our analysis, and the result is as follows.

\begin{Theorem} \label{thm:Te_MVES_unique}
%For $1/\sqrt{N-1} < r \leq 1$, any $\mathcal{T}' \in {\sf MVES}(\mathcal{R}(r))$ satisfies $\mathcal{T}' = \mathcal{T}_e$.
For any $1/\sqrt{N-1} < r \leq 1$,
we have ${\sf MVES}(\mathcal{R}(r))= \{ \mathcal{T}_e \}$;
i.e., there is only one MVES of $\mathcal{R}(r)$ for $1/\sqrt{N-1} < r \leq 1$ and that MVES is always given by the unit simplex.
%Consider $1/\sqrt{N-1} < r \leq 1$. If $\mathcal{T}_e' \in {\sf MVES}(\mathcal{R}(r))$, then $\mathcal{T}'_e = \mathcal{T}_e$.
\end{Theorem}
As an example, Fig.~\ref{fig:Rr_illus}.(b) is an instance where Theorem~\ref{thm:Te_MVES_unique} holds;
by visual observation of Fig.~\ref{fig:Rr_illus}.(b), we may argue that the MVES of $\mathcal{R}(r)$ for $N=3$ and $r > 1/\sqrt{2}$ should be the unit simplex.
Also, we should note that
the geometric problem in Theorem~\ref{thm:Te_MVES_unique} is interesting in its own right,
and the result could be of independent interest in other fields.
}

%The above theorems consider the case of three endmembers or more.
Before we finish this subsection, we should mention the case of $N=2$.
While the number of endmembers in practical scenarios is often a lot more than two, it is still interesting to know the identifiability for $N=2$.
\begin{Prop} \label{prop:N2}
Assume $N=2$. We have ${\sf MVES}(\mathcal{S}_L) = \{ \mathcal{T}_e \}$ if and only if the pure-pixel assumption holds.
\end{Prop}
We should recall that the pure-pixel assumption corresponds to $\gamma=1$.

\begin{figure}[htp!]
    \psfrag{e1}[Bc][Bc]{\huge$\bm e_1$}
    \psfrag{e2}[Bc][Bc]{\huge$\bm e_2$}
    \psfrag{e3}[Bc][Bc]{\huge$\bm e_3$}
    \psfrag{p12}[Bc][Bc]{\huge$\bm s_{n(2,3)}$}
    \psfrag{p13}[Bc][Bc]{\huge$\bm s_{n(3,2)}$}
    \psfrag{p21}[Bc][Bc]{\huge$\bm s_{n(1,3)}$}
    \psfrag{p23}[Bc][Bc]{\huge$\bm s_{n(3,1)}$}
    \psfrag{p31}[Bc][Bc]{\huge$\bm s_{n(1,2)}$}
    \psfrag{p32}[Bc][Bc]{\huge$\bm s_{n(2,1)}$}
    \psfrag{R}[Bc][Bl]{\huge$\setR (1/\sqrt{2})$}
    \centerline{\resizebox{0.4\textwidth}{!}{\includegraphics{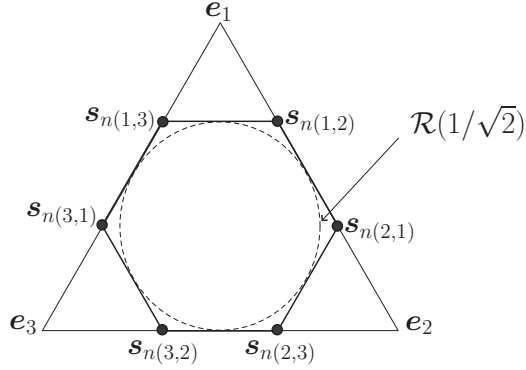}}}
    \caption{ Illustration of Assumption~\ref{assum:pure2}. $N=3$, $\alpha_{ij}= 2/3$ for all $i,j$.}
    \label{fig:pure2_illus}
\end{figure}

%----
{
\subsection{Further Discussion}

We have seen that the uniform pixel purity level $\gamma$ provides a key quantification on when MVES achieves perfect endmember identifiability.
Nevertheless, one may have these further questions:
How $\gamma$ is related to the abundance pixel set $\mathcal{S}_L$ exactly?
Can the relationship be characterized in an explicit and practically interpretable manner?
For example, as can be observed in the three-endmember illustrations in Fig.~\ref{fig:abundances},
satisfying the sufficient identifiability condition $\gamma > 1/\sqrt{N-1}$ in Theorem~\ref{thm:suff} seems to require some abundance pixels to lie on the boundary of $\mathcal{T}_e$.
However, from the definition of $\gamma$ in \eqref{eq:gamma_def},
it is not immediately clear how such a result can be deduced (e.g., how many pixels on the boundary, and which parts of the boundary?).
Unfortunately, explicit characterization of $\gamma$ w.r.t. $\mathcal{S}_L$ appears to be a difficult analysis problem.
In fact, even computing the value of $\gamma$ for a given $\mathcal{S}_L$ is generally an NP-hard problem\footnote{More accurately, verifying whether or not a convex body ($\mathcal{R}(r)$ here) belongs to a $\mathcal{V}$-polytope (${\rm conv} \mathcal{S}_L$ here) has been shown to be NP-hard~\cite{gritzmann1994complexity}.}~\cite{gritzmann1994complexity}.

Despite the aforementioned analysis bottleneck,
our empirical experience suggests that if every $\bm s_n$ follows a continuous distribution that has a support covering $\mathcal{R}(r)$ for $r> 1/{\sqrt{N-1}}$ (e.g., Dirichlet distributions),
and the number of pixels $L$ is large,
there is a large probability for MVES to achieve perfect identifiability.
The numerical results in Section~\ref{sec:sim} will confirm this.
Moreover, we can study special, but still meaningful, cases.
Herein we show one that uses the following assumption:
\begin{Assumption} \label{assum:pure2}
For every $i,j \in \{ 1,\ldots,N \}$, $i \neq j$,
there exists a pixel, whose index is denoted by $n(i,j)$,
such that its abundance vector takes the form
\begin{equation} \label{eq:s_pure2}
\bm s_{n(i,j)} = \alpha_{ij} \bm e_i + (1 - \alpha_{ij}) \bm e_j,
\end{equation}
for some coefficient $\alpha_{ij}$ that satisfies $\frac{1}{2} < \alpha_{ij} \leq 1$.
\end{Assumption}
Assumption~\ref{assum:pure2} means that we can find pixels that are constituted by two endmembers, with one dominating another as determined by the coefficient $\alpha_{ij} > \frac{1}{2}$.
Also, the pixels in \eqref{eq:s_pure2} lie on the edges of $\mathcal{T}_e$.
Fig.~\ref{fig:pure2_illus} gives an illustration for $N=3$.
Note that Assumption~\ref{assum:pure2} reduces to the pure-pixel assumption if $\alpha_{ij}= 1$ for all $i,j$.
Hence, Assumption~\ref{assum:pure2} may be seen as a more general assumption than the pure-pixel assumption.
In the example of $N=3$  in Fig.~\ref{fig:pure2_illus},
we see that $\gamma$ should increase as ${\alpha_{ij}}$'s increase.
In fact, this can be proven to be true for any $N \geq 2$.
%In this special case, a tractable lower bound on $\gamma$ can be proven.
\begin{Theorem} \label{thm:pure2}
Under Assumption~\ref{assum:pure2} and for $N \geq 2$, the uniform pixel purity level satisfies
\[ \gamma \geq \sqrt{ \frac{1}{N} \left[
\frac{(N\alpha-1)^2}{N-1} +1
\right] },
\]
where
\[ \alpha = \min_{\substack{ i,j \in \{ 1,\ldots,N \} \\ i \neq j }} \alpha_{ij} \]
is the smallest value of $\alpha_{ij}$'s.
\end{Theorem}
The proof of Theorem~\ref{thm:pure2} is given in Section~\ref{sec:proof:thm:pure2}.
Theorem~\ref{thm:pure2} is useful in the following way.
If we compare Theorems~\ref{thm:suff} and \ref{thm:pure2},
we see that the condition
\begin{equation*} \label{eq:con_pure2}
\sqrt{ \frac{1}{N} \left[
\frac{(N\alpha-1)^2}{N-1} +1
\right] } > \frac{1}{\sqrt{N-1}},
\end{equation*}
implies exact unique identifiability of MVES.
It is shown that
%\eqref{eq:con_pure2}
the above equation
is equivalent to
%$\alpha > 2/N$
\[ \alpha > \frac{2}{N}, \]
for $N \geq 3$.
By also noting $\frac{1}{2} < \alpha \leq 1$ in Assumption~\ref{assum:pure2},
and the fact that $\frac{1}{2} \geq \frac{2}{N}$ for $N \geq 4$,
we have the following conclusion.
\begin{Corollary} \label{cor:pure2_id}
Suppose that Assumption~\ref{assum:pure2} holds.
For $N=3$,
the exact unique identifiability condition ${\sf MVES}(\mathcal{S}_L) = \{ \mathcal{T}_e \}$ is achieved if $\alpha_{ij} > \frac{2}{3}$ for all $i,j$.
For $N \geq 4$,
the condition ${\sf MVES}(\mathcal{S}_L) = \{ \mathcal{T}_e \}$ is always achieved
 %for any $\alpha_{ij}$'s
 (subject to $\frac{1}{2} < \alpha_{ij} \leq 1$ in Assumption~\ref{assum:pure2}).
\end{Corollary}
%It is worthwhile to notice the implication of Corollary~\ref{cor:pure2_id} for the case of $N \geq 4$---namely, MVES always provides perfect identifiability under Assumption~\ref{assum:pure2}.
The implication of Corollary~\ref{cor:pure2_id} is particularly interesting for $N \geq 4$---MVES for $N \geq 4$ always provides perfect identifiability under Assumption~\ref{assum:pure2}.
%This suggests that MVES may be able to deal with heavily mixed cases for larger $N$.
However, we should also note that this result is under the premise of Assumption~\ref{assum:pure2}.
In particular,
%at least $N(N-1)$ pixels are required to satisfy Assumption~\ref{assum:pure2}, and
it is seen that to satisfy Assumption~\ref{assum:pure2} for general $\alpha_{ij}$'s, the number of pixels $L$ should be no less than $N(N-1)$.
This
%somehow
implies that we would need more pixels to achieve perfect MVES identifiability as $N$ increases.

We finish with mentioning some arising open problems.
From the above discussion, it is natural to further question whether \eqref{eq:s_pure2} in Assumption~\ref{assum:pure2} can be relaxed to combinations of three endmembers, or more.
Also, the whole work has so far assumed the noiseless case, and sensitivity in the noisy case has not been touched.
These challenges are left as future work.
}

\section{Proof of The Main Results}
\label{sec:main_proof}

This section provides the proof of the main results described in the previous section.
Readers who are more interested in numerical experiments may jump to Section~\ref{sec:sim}.

\subsection{Proof of Proposition~\ref{prop:mut}}

The following lemma will be used to prove Proposition~\ref{prop:mut}:
\begin{Lemma} \label{lem:vol_trans}
Let %$f: \mathbb{R}^N \rightarrow \mathbb{R}^M$ takes the form
$f(\bm x) = \bm A \bm x$, where $\bm A \in \mathbb{R}^{M \times N}$, $M \geq N$,
and suppose that $\bm A$ has full column rank.
\begin{itemize}
\item[(a)] Let $\mathcal{T}_G \subset \mathbb{R}^N$ be an $(N-1)$-dimensional simplex,
and suppose $\mathcal{T}_G \subset {\rm aff} \{ \bm e_1, \ldots, \bm e_N \}$.
We have
\begin{equation} \label{eq:vol1}
{\rm vol}( f(\mathcal{T}_G) ) = \alpha
 \cdot {\rm vol}(\mathcal{T}_G),
\end{equation}
where
$\alpha = \sqrt{\frac{ \det( \bar{\bm A}^T \bar{\bm A} ) }{N} }$, and
$ \bar{\bm A}= [~ \bm a_1 - \bm a_N, \bm a_2 - \bm a_N, \ldots, \bm a_{N-1} - \bm a_N ~]$.
Also, it holds true that
%\begin{equation} \label{eq:aff1}
$f(\mathcal{T}_G) \subset {\rm aff}\{ \bm a_1, \ldots, \bm a_N \}.$
%\end{equation}

\item[(b)] Let $\mathcal{T}_H \subset \mathbb{R}^M$ be an $(N-1)$-dimensional simplex,
and suppose $\mathcal{T}_H \subset {\rm aff} \{ \bm a_1, \ldots, \bm a_N \}$.
We have
\begin{equation} \label{eq:vol2}
{\rm vol}( f^{-1}(\mathcal{T}_H) ) =
\frac{1}{\alpha} \cdot {\rm vol}(\mathcal{T}_H),
\end{equation}
and
%\begin{equation} \label{eq:aff2}
$f^{-1}(\mathcal{T}_H) \subset {\rm aff}\{ \bm e_1, \ldots, \bm e_N \}.$
%\end{equation}
\end{itemize}
\end{Lemma}

The proof of Lemma~\ref{lem:vol_trans} is relegated to Appendix~\ref{proofsec:lem:vol_trans}.
Now, suppose that ${\sf MVES}( \mathcal{S}_L ) = \{ \mathcal{T}_e \}$,
but ${\sf MVES}( \mathcal{X}_L ) \neq \{ \mathcal{T}_a \}$.
Let $\mathcal{T}_H$ be an MVES of $\mathcal{X}_L$.
By the MVES definition (see Definition~\ref{def:mves}), we have
\begin{equation} \label{eq:mves_X_cond}
\begin{aligned}
& \mathcal{X}_L  \subseteq \mathcal{T}_H, \quad \mathcal{T}_H \subseteq {\rm aff}\{ \bm x_1, \ldots, \bm x_L \}, \\
& {\rm vol}(\mathcal{T}_H)  \leq {\rm vol}(\mathcal{T}_a).
\end{aligned}
\end{equation}
Recall that $[ \bm s_1, \ldots, \bm s_L ~]$ is assumed to have full row rank and satisfy $\bm 1^T \bm s_n= 1$ for all $n$. From these assumptions, one can prove that ${\rm aff}\{ \bm s_1, \ldots, \bm s_L \} = {\rm aff}\{ \bm e_1, \ldots, \bm e_N \}$, and ${\rm aff}\{ \bm x_1, \ldots, \bm x_L \} = {\rm aff}\{ \bm a_1, \ldots, \bm a_N \}$; see \cite[Lemma~1]{Chan2008} for example.
Then, by applying Lemma~\ref{lem:vol_trans}.(b) to \eqref{eq:mves_X_cond}, we obtain
\begin{align*}
& \mathcal{S}_L  \subseteq f^{-1}( \mathcal{T}_H ), \quad f^{-1}( \mathcal{T}_H ) \subseteq {\rm aff}\{ \bm e_1, \ldots, \bm e_N \}, \\
& {\rm vol}(f^{-1}(\mathcal{T}_H)) \leq {\rm vol}(f^{-1}(\mathcal{T}_a)) = {\rm vol}(\mathcal{T}_e).
\end{align*}
The above equation implies that $\mathcal{T}_e$ is not the only MVES of $\mathcal{S}_L$, which is a contradiction.

On the other hand, suppose that ${\sf MVES}( \mathcal{X}_L ) = \{ \mathcal{T}_a \}$, but ${\sf MVES}( \mathcal{S}_L ) \neq \{ \mathcal{T}_e \}$.
This statement can be shown to be a contradiction, by the same proof as above (particularly, the incorporation of Lemma~\ref{lem:vol_trans}.(a)).
The proof of Proposition~\ref{prop:mut} is therefore complete.

\subsection{Proof of Theorem~\ref{thm:nec}}

The proof is done by contradiction.
Suppose that ${\sf MVES}(\mathcal{S}_L) = \{ \mathcal{T}_e \}$, but $\rho \leq \frac{1}{\sqrt{N-1}}$.
%We will show that the above statement is a contradiction.
%To this end, let
Recall
\begin{equation} \label{eq:Rr}
\mathcal{R}(r) = \mathcal{T}_e \cap \{ \bm s \in \mathbb{R}^N ~|~ \| \bm s \| \leq r \}.
\end{equation}
The proof is divided into four steps.

{\em Step 1:} \
We show that any $\mathcal{V} \in {\sf MVES}(\mathcal{R}(\rho))$ is also an MVES of $\mathcal{S}_L$.
%Since every $\bm s_i$ satisfies $\bm s_i \in \mathcal{T}_e$ and $\| \bm s_i \| \leq \rho$, we have
To prove it, note that
\begin{equation} \label{eq:SL_Rr}
\mathcal{S}_L \subseteq \mathcal{R}(\rho).
\end{equation}
Eq.~\eqref{eq:SL_Rr} implies that
\begin{equation} \label{eq:volUV}
{\rm vol}(\mathcal{U}) \leq {\rm vol}(\mathcal{V}), \text{~for all $\mathcal{U} \in {\sf MVES}(\mathcal{S}_L)$, $\mathcal{V} \in {\sf MVES}(\mathcal{R}(\rho))$}.
\end{equation}
Also, since $\mathcal{T}_e$ encloses $\mathcal{R}(\rho)$, we have
\begin{equation} \label{eq:volVTe}
{\rm vol}(\mathcal{V}) \leq {\rm vol}(\mathcal{T}_e), \text{~for all $\mathcal{V} \in {\sf MVES}(\mathcal{R}(\rho))$}.
\end{equation}
Since we assume ${\sf MVES}(\mathcal{S}_L) = \{ \mathcal{T}_e \}$ in the beginning,
we observe from \eqref{eq:volUV} and \eqref{eq:volVTe} that ${\rm vol}(\mathcal{U})= {\rm vol}(\mathcal{V})$ for all $\mathcal{U} \in {\sf MVES}(\mathcal{S}_L)$, $\mathcal{V} \in {\sf MVES}(\mathcal{R}(\rho))$.
The above equality, together with \eqref{eq:SL_Rr}, implies that any $\mathcal{V} \in {\sf MVES}(\mathcal{R}(\rho))$ is an MVES of $\mathcal{S}_L$ (or satisfies $\mathcal{V} \in {\sf MVES}(\mathcal{S}_L))$.

{\em Step 2:} \ We give an alternative representation of $(N-1)$-dimensional simplices on ${\rm aff}\{ \bm e_1, \ldots, \bm e_N \}$, which will facilitate the proof.
The affine hull ${\rm aff}\{ \bm e_1, \ldots, \bm e_N \}$ can be equivalently expressed as
\begin{equation} \label{eq:aff_e_equiv}
{\rm aff}\{ \bm e_1, \ldots, \bm e_N \} = \{ \bm s= \bm C \bm \theta + \bm d ~|~ \bm \theta \in \mathbb{R}^{N-1} \},
\end{equation}
where
\begin{equation*}
\bm d = \frac{1}{N} \sum_{i=1}^N \bm e_i = \frac{1}{N} \bm 1,
\end{equation*}
and $\bm C \in \mathbb{R}^{N \times (N-1)}$ is the first $N-1$ principal left singular vectors of
$\bm R = [~ \bm e_1 - \bm d, \ldots, \bm e_N - \bm d ~]$; see \cite{Chan2008,Chan2009}.
We note that
\begin{equation*}
\bm R = \bm I - \frac{1}{N} \bm 1 \bm 1^T,
\end{equation*}
which, as a standard matrix result, its first $N-1$ principal left singular vector can be shown to be any $\bm C$ such that
\begin{equation} \label{eq:Cd}
\bm U = \left[~ \bm C, ~ \frac{1}{\sqrt{N}} \bm 1 ~\right]
\end{equation}
is a unitary matrix. Or, equivalently, $\bm C$ is any semi-unitary matrix such that $\bm C^T \bm d = \bm 0$.

Recall that an $(N-1)$-dimensional simplex $\mathcal{V} \subseteq {\rm aff}\{ \bm e_1, \ldots, \bm e_N \}$ can be written as
\begin{equation*}
\mathcal{V} = {\rm conv}\{ \bm v_1, \ldots, \bm v_N \},
\end{equation*}
where $\bm v_i \in {\rm aff}\{ \bm e_1, \ldots, \bm e_N \}$ for all $i$.
By \eqref{eq:aff_e_equiv},
each $\bm v_i \in {\rm aff}\{ \bm e_1, \ldots, \bm e_N \}$
can be represented by $\bm v_i = \bm C \bm w_i + \bm d$ for some $\bm w_i \in \mathbb{R}^{N-1}$.
Applying this result to ${\rm conv}\{ \bm v_1, \ldots, \bm v_N \}$,
%one can show that
%\begin{equation} \label{eq:V_alt}
%\mathcal{V} = \bigg\{ \bm x = \bm C \bm \theta + \bm d ~\bigg|~  \bm \theta = \sum_{i=1}^N \alpha_i \bm w_i, \bm \alpha \geq \bm 0, \bm 1^T \bm \alpha = 1 \bigg\},
%\end{equation}
we obtain the following equivalent representation of $\mathcal{V}$
\begin{equation} \label{eq:V_alt}
\mathcal{V} = \{ \bm s = \bm C \bm \theta + \bm d ~|~  \bm \theta \in \mathcal{W} \},
\end{equation}
where
\begin{equation} \label{eq:W_alt}
\mathcal{W} = {\rm conv}\{ \bm w_1, \ldots, \bm w_N \}.
\end{equation}
Also, by the simplex volume formula \eqref{eq:vol_formula} and the semi-unitarity of $\bm C$, the following relation is shown
\begin{equation} \label{eq:vol_eq}
{\rm vol}(\mathcal{V}) = {\rm vol}(\mathcal{W}).
\end{equation}

{\em Step 3:} \ We show that there are infinitely many MVES of $\mathcal{R}(\rho)$ for $\frac{1}{\sqrt{N}} <  \rho \leq \frac{1}{\sqrt{N-1}}$.
Consider the following lemma.
\begin{Lemma} \label{lem:Rr_eq_Cr}
Let
\begin{align}
%\mathcal{R}(r) & = \mathcal{C}(r), \\
\mathcal{C}(r) & = {\rm aff}\{ \bm e_1, \ldots, \bm e_N \} \cap \{ \bm s \in \mathbb{R}^N ~|~ \| \bm s \| \leq r \}.
\end{align}
denote a $2$-norm ball on ${\rm aff}\{ \bm e_1, \ldots, \bm e_N \}$.
If $\frac{1}{\sqrt{N}} <  r \leq \frac{1}{\sqrt{N-1}}$,
then $\mathcal{R}(r)$ in \eqref{eq:Rr} equals $\mathcal{C}(r)$.
\end{Lemma}

{\it Proof of Lemma~\ref{lem:Rr_eq_Cr}:} \
Note that $\mathcal{R}(r) \subseteq \mathcal{C}(r)$.
Hence, to prove Lemma~\ref{lem:Rr_eq_Cr}, it suffices to show that $\mathcal{C}(r) \subseteq \mathcal{R}(r)$.
%To prove the first part of Lemma 3, we seek to show that any $\bm x \in \mathcal{C}(r)$ satisfies $\bm x \in \mathcal{R}(r)$.
By the equivalent affine hull representation in \eqref{eq:aff_e_equiv},
we can write
$\mathcal{C}(r)= \{ \bm s= \bm C \bm \theta + \bm d ~|~  \| \bm s \| \leq r \}$.
By substituting $\bm s= \bm C \bm \theta + \bm d$ into $\| \bm s \| \leq r$,
we get, for any $\bm s \in \mathcal{C}(r)$,
\begin{subequations}
\begin{align}
% &  \| \bm s \|^2 \leq r^2 \label{eq:ta} \\
\| \bm s \|^2 \leq r^2 \Longleftrightarrow & \| \bm \theta \|^2 + \| \bm d \|^2  \leq r^2 \label{eq:tb} \\
\Longleftrightarrow & \| \bm \theta \|^2  \leq r^2 - \frac{1}{N}, \label{eq:tc}
\end{align}
\end{subequations}
where \eqref{eq:tb} is obtained by using the orthogonality in \eqref{eq:Cd}; \eqref{eq:tc} is by $\| \bm d \|^2 = \frac{1}{N}$.
Hence, $\mathcal{C}(r)$ can be rewritten as
\begin{equation} \label{eq:Cr_alt}
\mathcal{C}(r) = \{ \bm s = \bm C \bm \theta + \bm d ~|~ \| \bm \theta \|^2 \leq r^2 - 1/N \}.
\end{equation}
Moreover, by letting $\bm c^i$ and $\bm u^i$ denote the $i$th rows of $\bm C$ and $\bm U$ respectively, we have
\begin{subequations}\label{eq:u}
\begin{align}
s_i & = [ \bm c^i ]^T \bm \theta + d_i \label{eq:ua} \\
& \geq - \| \bm c^i \| \| \bm \theta \| + \frac{1}{N}  \label{eq:ub} \\
& \geq - \sqrt{ \frac{N-1}{N} } \cdot \sqrt{\frac{1}{(N-1) \cdot N}} + \frac{1}{N} = 0,  \label{eq:uc}
%& = [ \bm q^i ]^T \begin{bmatrix} \bm \theta \\ 0 \end{bmatrix} + \frac{1}{N} \\
%& \geq  - \| \bm q^i \| \| \bm \theta \| + \frac{1}{N} \\
%& \geq - \frac{1}{(N-1) \cdot N} + \frac{1}{N} \geq 0.
\end{align}
\end{subequations}
where \eqref{eq:ub} is due to the Cauchy-Schwartz inequality;
\eqref{eq:uc} is due to \eqref{eq:tc}, $r \leq \frac{1}{\sqrt{N-1}}$, and the fact that $1 = \| \bm u^i \|^2 = \frac{1}{N} + \| \bm c^i \|^2$ (see \eqref{eq:Cd} and note its orthogonality).
Eq.~\eqref{eq:u} suggests that any $\bm s \in \mathcal{C}(r)$ automatically satisfies $\bm s \geq \bm 0$, and hence, $\bm s \in \mathcal{R}(r)$.
We therefore conclude that $\mathcal{C}(r) = \mathcal{R}(r)$.
\hfill $\blacksquare$

By Lemma~\ref{lem:Rr_eq_Cr}, we can replace $\mathcal{R}(\rho)$ by $\mathcal{C}(\rho)$ and consider the MVES of the latter.
Suppose that $\mathcal{V} \in {\sf MVES}(\mathcal{C}(\rho))$.
Our argument is that a suitably rotated version of $\mathcal{V}$ is also an MVES of $\mathcal{C}(\rho)$.
To be precise, use the representation in \eqref{eq:V_alt}-\eqref{eq:W_alt} to describe $\mathcal{V}$.
Comparing \eqref{eq:V_alt}-\eqref{eq:W_alt} and \eqref{eq:Cr_alt}, we see that $\mathcal{C}(\rho) \subseteq \mathcal{V}$ is equivalent to
\begin{equation} \label{eq:WsubseteqB}
\{ \bm \theta ~|~ \| \bm \theta \|^2 \leq \rho^2 - 1/N \} \subseteq \mathcal{W}.
\end{equation}
From $\mathcal{W}$, let us construct another simplex
\begin{equation}
\mathcal{V}' = \{ \bm s = \bm C \bm Q  \bm \theta + \bm d ~|~ \bm \theta \in \mathcal{W} \},
\end{equation}
where $\bm Q \in \mathbb{R}^{(N-1) \times (N-1)}$ is a unitary matrix.
Due to \eqref{eq:WsubseteqB}, $\mathcal{V}'$ can be verified to satisfy $\mathcal{C}(\rho) \subseteq \mathcal{V}'$.
Also, by observing the semi-unitarity of $\bm C \bm Q$, the volume of $\mathcal{V}'$ is shown to equal
\begin{equation*}
{\rm vol}(\mathcal{V}') = {\rm vol}(\mathcal{W}) = {\rm vol}(\mathcal{V}).
\end{equation*}
In other words, $\mathcal{V}'$ is also an MVES of $\mathcal{C}(\rho)$.
In fact, the argument above holds for any unitary $\bm Q$.
%thereby resulting in infinitely many MVES of $\mathcal{C}(\rho)$.
Since there are infinitely many unitary $\bm Q$ for $N \geq 3$ (note that $\bm Q \in \mathbb{R}^{(N-1) \times (N-1)}$), we also have infinitely many MVESs of $\mathcal{C}(\rho)$ for $N \geq 3$.

{\it Step 4:} \
We combine the results in the above steps to draw conclusion.
Step 1 shows that any $\mathcal{V} \in {\sf MVES}(\mathcal{R}(\rho))$ is also an MVES of $\mathcal{S}_L$,
while Step 3 shows that $\mathcal{R}(\rho)$ has infinitely many MVESs for $\rho \leq \frac{1}{\sqrt{N-1}}$, $N \geq 3$.
This contradicts the assumption that there is only one MVES of $\mathcal{S}_L$.
The proof of Theorem~\ref{thm:nec} is therefore complete.

%
%\begin{Lemma} \label{lem:nonunique_Cr}
%For $r> \frac{1}{\sqrt{N}}$, $\mathcal{C}(r)$ has infinitely many MVES.
%To be precise, let
%\begin{equation}
%f(\bm s) = \bm C \bm Q \bm C^T (\bm s - \bm d) + \bm d,
%\label{eq:fs_Cr}
%\end{equation}
%where $\bm d = \frac{1}{N} \bm 1$, $\bm C \in \mathbb{R}^{N \times (N-1)}$ is any semi-unitary matrix such that $[~ \bm C, \frac{1}{\sqrt{N}} \bm 1 ~]$ is unitary, and $\bm Q \in \mathbb{R}^{(N-1) \times (N-1)}$.
%If $\mathcal{V} \in {\sf MVES}(\mathcal{C}(r))$, then $f(\mathcal{V}) \in {\sf MVES}(\mathcal{C}(r))$ for any unitary $\bm Q$.
%\end{Lemma}
%
%The proofs of Lemmas~\ref{lem:Rr_eq_Cr} and \ref{lem:nonunique_Cr} are shown in ??? and ????, respectively.
%By Lemmas~\ref{lem:Rr_eq_Cr} and \ref{lem:nonunique_Cr}, there are infinitely many $\mathcal{V} \in {\sf MVES}(\mathcal{R}(\rho))$ for $\frac{1}{\sqrt{N}} <  \rho \leq \frac{1}{\sqrt{N-1}}$, which contradicts the assumption ${\sf MVES}(\mathcal{S}_L) = \{ \mathcal{T}_e \}$.
%We therefore conclude that Theorem~\ref{thm:nec} is true.

\subsection{Proof of Theorem~\ref{thm:suff}}

To facilitate our proof, let us introduce the following fact.

\begin{Fact} \label{fact:CD}
Let $\mathcal{C}, \mathcal{D} \subseteq \mathbb{R}^n$ be two sets of identical dimension, with $\mathcal{C} \subseteq \mathcal{D}$.
If $\mathcal{D} \subseteq \mathcal{T}$ for some $\mathcal{T} \in {\sf MVES}(\mathcal{C})$, then $\mathcal{T} \in {\sf MVES}(\mathcal{D})$ and ${\sf MVES}(\mathcal{D}) \subseteq {\sf MVES}(\mathcal{C})$.
\end{Fact}

{\it Proof of Fact~\ref{fact:CD}:} \
Note that $\mathcal{C} \subseteq \mathcal{D}$ implies that any $\mathcal{T}' \in {\sf MVES}(\mathcal{D})$
%satisfies $\mathcal{C} \subseteq \mathcal{T}'$.
is a simplex enclosing $\mathcal{C}$.
Since $\mathcal{T}$ is a minimum volume simplex among all the $\mathcal{C}$-enclosing simplices, we have
\begin{equation} \label{eq:CD_fact}
{\rm vol}(\mathcal{T}) \leq {\rm vol}(\mathcal{T}') ~ \text{for all $\mathcal{T}'\in {\sf MVES}(\mathcal{D})$}.
\end{equation}
Moreover, the condition $\mathcal{D} \subseteq \mathcal{T}$ implies that $\mathcal{T}$ is also a $\mathcal{D}$-enclosing simplex, and, as a result, equality in \eqref{eq:CD_fact} holds.
It also follows that any $\mathcal{T}' \in {\sf MVES}(\mathcal{D})$ is also an MVES of $\mathcal{C}$.
\hfill $\blacksquare$

\medskip
Now we proceed with the main proof.

{\it Step 1:} \ We show that
\begin{equation} \label{eq:proof_suff_t0}
\mathcal{T}_e \in {\sf MVES}(\mathcal{R}(r)),
%\text{~for any $\frac{1}{\sqrt{N-1}} < r \leq 1$.}
\text{~for any $r \geq \frac{1}{\sqrt{N-1}}$.}
\end{equation}
Note from the definition of $\mathcal{R}(r)$ in \eqref{eq:Rr_def} that
\begin{equation} \label{eq:proof_suff_t1}
\mathcal{C}\left( \tfrac{1}{\sqrt{N-1}} \right) = \mathcal{R}\left( \tfrac{1}{\sqrt{N-1}} \right) \subseteq \mathcal{R}(r) \subseteq \mathcal{T}_e,
\end{equation}
for any $r \in [ 1/\sqrt{N-1}, 1]$, where the first equality is by  Lemma~\ref{lem:Rr_eq_Cr}. We prove that

\begin{Lemma} \label{lem:Te_MVES}
The unit simplex $\mathcal{T}_e$ is an MVES of $\mathcal{C}(1/\sqrt{N-1})$.
\end{Lemma}

The proof of Lemma~\ref{lem:Te_MVES} is relegated to Appendix~\ref{proofsec:lem:Te_MVES}.
By applying Fact~\ref{fact:CD} and Lemma~\ref{lem:Te_MVES}  to \eqref{eq:proof_suff_t1}, we obtain $\mathcal{T}_e \in {\sf MVES}(\mathcal{R}(r))$ for $r \in [ 1/\sqrt{N-1}, 1]$.

{\it Step 2:} \ We prove that
\begin{equation} \label{eq:proof_suff_t15}
{\sf MVES}( \mathcal{S}_L ) \subseteq {\sf MVES}( \mathcal{R}(\gamma) ), \text{~for $\gamma \geq \frac{1}{\sqrt{N-1}}$.}
\end{equation}
By the definition of $\gamma$ in \eqref{eq:gamma_def}, we have
\begin{equation} \label{eq:proof_suff_t2}
\mathcal{R}(\gamma) \subseteq {\rm conv}\mathcal{S}_L \subseteq \mathcal{T}_e.
\end{equation}
Also, in Step~1, it has been identified that $\mathcal{T}_e \in {\sf MVES}(\mathcal{R}(r))$ for $r \in [ 1/\sqrt{N-1}, 1]$.
Hence, for $\gamma \geq 1/\sqrt{N-1}$, we can apply Fact~\ref{fact:CD} to \eqref{eq:proof_suff_t2} to obtain
\begin{equation}  \label{eq:proof_suff_t3}
{\sf MVES}( {\rm conv}\mathcal{S}_L ) \subseteq {\sf MVES}( \mathcal{R}(\gamma) ).
\end{equation}
Next, we use a straightforward fact in convex analysis: for a convex set $\mathcal{T}$,
the condition $\mathcal{C} \subset \mathcal{T}$ is the same as ${\rm conv}\mathcal{C} \subset \mathcal{T}$, and vice versa.
In the context here, this implies that any MVES of ${\rm conv}\mathcal{S}_L$ also encloses $\mathcal{S}_L$, and the converse is also true.
Hence, we have
\begin{equation} \label{eq:proof_suff_t4}
{\sf MVES}( {\rm conv}\mathcal{S}_L ) = {\sf MVES}( \mathcal{S}_L ).
\end{equation}
By combining \eqref{eq:proof_suff_t3} and \eqref{eq:proof_suff_t4}, Eq.~\eqref{eq:proof_suff_t15} is obtained.

{\it Step 3:} \ We prove that
\begin{equation} \label{eq:proof_suff_t5}
{\sf MVES}( \mathcal{R}(\gamma) ) = \{ \mathcal{T}_e \}, \text{~for $\gamma > \frac{1}{\sqrt{N-1}}$.}
\end{equation}
It has been shown in
%Lemma~\ref{lem:Te_MVES}
Step 1
that $\mathcal{T}_e \in {\sf MVES}( \mathcal{R}(\gamma) )$.
The question is whether there exists another MVES $\mathcal{T}' \in {\sf MVES}( \mathcal{R}(\gamma) )$, with $\mathcal{T}' \neq \mathcal{T}_e$.
{By Theorem~\ref{thm:Te_MVES_unique}, such a $\mathcal{T}'$ does not exist.
Thus, \eqref{eq:proof_suff_t5} is obtained.
}
%It turns out that such a $\mathcal{T}'$ does not exist, as we prove in the following theorem.
%\begin{Theorem} \label{thm:Te_MVES_unique}
%%For $1/\sqrt{N-1} < r \leq 1$, any $\mathcal{T}' \in {\sf MVES}(\mathcal{R}(r))$ satisfies $\mathcal{T}' = \mathcal{T}_e$.
%For any $1/\sqrt{N-1} < r \leq 1$,
%we have ${\sf MVES}(\mathcal{R}(r))= \{ \mathcal{T}_e \}$;
%i.e., there is only one MVES of $\mathcal{R}(r)$ for $1/\sqrt{N-1} < r \leq 1$ and that MVES is always given by the unit simplex.
%%Consider $1/\sqrt{N-1} < r \leq 1$. If $\mathcal{T}_e' \in {\sf MVES}(\mathcal{R}(r))$, then $\mathcal{T}'_e = \mathcal{T}_e$.
%\end{Theorem}
%Note that Theorem~\ref{thm:Te_MVES_unique} shows a unique MVES condition for $\mathcal{R}(r)$, which is interesting in its own right and could be of independent interest in other fields.
%The proof of Theorem~\ref{thm:Te_MVES_unique} is shown in the next subsection.
%By
%%Lemma~\ref{lem:Te_MVES} and
%Theorem~\ref{thm:Te_MVES_unique}, \eqref{eq:proof_suff_t5} is obtained.

{\it Step 4:} \ We combine the results in Steps 2 and 3.
Specifically, by \eqref{eq:proof_suff_t15} and \eqref{eq:proof_suff_t5}, we get ${\sf MVES}( \mathcal{S}_L ) \subseteq \{ \mathcal{T}_e \}$.
As $\mathcal{S}_L$ is enclosed by $\mathcal{T}_e$, we further deduce ${\sf MVES}( \mathcal{S}_L ) = \{ \mathcal{T}_e \}$.
Theorem~\ref{thm:suff} is therefore proven.

\subsection{Proof of Theorem~\ref{thm:Te_MVES_unique}}

Let $\mathcal{T}' \in {\sf MVES}(\mathcal{R}(r))$ be an arbitrary MVES of $\mathcal{R}(r)$ for $1/\sqrt{N-1} < r \leq 1$.
We prove Theorem~\ref{thm:Te_MVES_unique} by showing that $\mathcal{T}' = \mathcal{T}_e$ is always true.
The proof is divided into three steps.

{\it Step 1:} \
We show that $$\mathcal{T}' \in {\sf MVES}(\mathcal{R}( 1/\sqrt{N-1} )).$$
%It is proven in \eqref{eq:proof_suff_t0} that $\mathcal{T}_e \in {\sf MVES}(\mathcal{R}(r))$ for all $1/\sqrt{N-1} \leq r \leq 1$.
%Since both $\mathcal{T}_e, \mathcal{T}'$ are MVES of $\mathcal{R}(r)$ for $r > 1/\sqrt{N-1}$, we have
%\begin{equation*}
%{\rm vol}(\mathcal{T}_e) = {\rm vol}(\mathcal{T}').
%\end{equation*}
%Also, by noting that
%\begin{equation*}
%\mathcal{R}( 1/\sqrt{N-1} ) \subseteq \mathcal{R}(r) \subseteq \mathcal{T}'
%\end{equation*}
%for $r > 1/\sqrt{N-1}$, which means that $\mathcal{T}'$ encloses $\mathcal{R}( 1/\sqrt{N-1} )$, we conclude that $\mathcal{T}' \in {\sf MVES}(\mathcal{R}( 1/\sqrt{N-1} ))$.
To prove this, note that $\mathcal{R}( 1/\sqrt{N-1} ) \subseteq \mathcal{R}( r )$ for all  $1/\sqrt{N-1} \leq r \leq 1$.
Also, it has been shown in \eqref{eq:proof_suff_t0} that $\mathcal{T}_e \in {\sf MVES}(\mathcal{R}(r))$ for all $1/\sqrt{N-1} \leq r \leq 1$.
Applying Fact~\ref{fact:CD} to the above two results yields
$${\sf MVES}(\mathcal{R}( r )) \subseteq {\sf MVES}(\mathcal{R}( 1/\sqrt{N-1} )), %~ \text{for all $1/\sqrt{N-1} \leq r \leq 1$}.
$$
for all $1/\sqrt{N-1} \leq r \leq 1$.
Since $\mathcal{T}' \in {\sf MVES}(\mathcal{R}(r))$ for $1/\sqrt{N-1} < r \leq 1$,
it follows that $\mathcal{T}' \in {\sf MVES}(\mathcal{R}(1/\sqrt{N-1}))$ is also true.

{\it Step 2:} \
To proceed further, we apply the equivalent representation in \eqref{eq:V_alt}-\eqref{eq:W_alt} to rewrite $\mathcal{T}_e$ as
\begin{equation} \label{eq:Te_alt}
\mathcal{T}_e = \{ \bm s = \bm C \bm \theta + \bm d ~|~ \bm \theta \in \mathcal{W}_e \}
\end{equation}
%where $\mathcal{W}_e = {\rm conv}\{ \bm w_{e,1},\ldots, \bm w_{e,N} \} \subseteq \mathbb{R}^{N-1}$ for some $\bm w_{e,1},\ldots, \bm w_{e,N}$.
for some $(N-1)$-dimensional simplex $\mathcal{W}_e \subseteq \mathbb{R}^{N-1}$.
Similarly, we can characterize $\mathcal{T}'$ by
\begin{equation} \label{eq:T'_alt}
\mathcal{T}' = \{ \bm s = \bm C \bm \theta + \bm d ~|~ \bm \theta \in \mathcal{W}' \}
\end{equation}
%where $\mathcal{W}' = {\rm conv}\{ \bm w_{1}',\ldots, \bm w_{N}' \} \subseteq \mathbb{R}^{N-1}$ for some $\bm w_{1}',\ldots, \bm w_{N}'$.
for some $(N-1)$-dimensional simplex $\mathcal{W}' \subseteq \mathbb{R}^{N-1}$.
Also, by noting $\mathcal{R}(r) = \mathcal{T}_e \cap \mathcal{C}(r)$, the expression of $\mathcal{C}(r)$ in \eqref{eq:Cr_alt},
and $\mathcal{R}(r) = \mathcal{C}(r)$ for $r =1/\sqrt{N-1}$ (see Lemma~\ref{lem:Rr_eq_Cr}),
$\mathcal{R}(r)$ can be expressed as
\ifconfver
    \begin{equation} \label{eq:Rr_alt}
    \begin{aligned}
    & \mathcal{R}(r) = \\
    & \left\{ \begin{array}{ll}
    \{ \bm s = \bm C \bm \theta + \bm d ~|~ \bm \theta \in \mathcal{B}(\sqrt{r^2 -1/N}) \}, & r = \frac{1}{\sqrt{N-1}} \\
    \{ \bm s = \bm C \bm \theta + \bm d ~|~ \bm \theta \in \mathcal{W}_e \cap \mathcal{B}(\sqrt{r^2 -1/N}) \}, & r > \frac{1}{\sqrt{N-1}}
    \end{array} \right.
    \end{aligned}
    \end{equation}
\else
    \begin{equation} \label{eq:Rr_alt}
    \mathcal{R}(r) =
    \left\{ \begin{array}{ll}
    \{ \bm s = \bm C \bm \theta + \bm d ~|~ \bm \theta \in \mathcal{B}(\sqrt{r^2 -1/N}) \}, & r = \frac{1}{\sqrt{N-1}} \\
    \{ \bm s = \bm C \bm \theta + \bm d ~|~ \bm \theta \in \mathcal{W}_e \cap \mathcal{B}(\sqrt{r^2 -1/N}) \}, & r > \frac{1}{\sqrt{N-1}}
    \end{array} \right.
    \end{equation}
\fi
where
\begin{equation} \label{eq:Br_ball}
\mathcal{B}(r) = \{ \bm \theta \in \mathbb{R}^{N-1} ~|~  \| \bm \theta \| \leq r \}.
\end{equation}
Now, by comparing \eqref{eq:T'_alt}-\eqref{eq:Rr_alt}, the following result can be proven:
\ifconfver
    \begin{equation} \label{eq:T_W_equiv}
    \begin{aligned}
    & \mathcal{T}' \in {\sf MVES}(\mathcal{R}(r)) \Longleftrightarrow  \\
    & \mathcal{W}' \in
    \left\{ \begin{array}{ll}
    {\sf MVES}\left(  \mathcal{B}(\sqrt{r^2 -1/N}) \right), & r = \frac{1}{\sqrt{N-1}} \\
    {\sf MVES}\left(\mathcal{W}_e \cap \mathcal{B}(\sqrt{r^2 -1/N}) \right), & r > \frac{1}{\sqrt{N-1}}
    \end{array} \right.
    \end{aligned}
    \end{equation}
\else
    \begin{equation} \label{eq:T_W_equiv}
    \mathcal{T}' \in {\sf MVES}(\mathcal{R}(r)) \Longleftrightarrow \mathcal{W}' \in
    \left\{ \begin{array}{ll}
    {\sf MVES}\left(  \mathcal{B}(\sqrt{r^2 -1/N}) \right), & r = \frac{1}{\sqrt{N-1}} \\
    {\sf MVES}\left(\mathcal{W}_e \cap \mathcal{B}(\sqrt{r^2 -1/N}) \right), & r > \frac{1}{\sqrt{N-1}}
    \end{array} \right.
    \end{equation}
\fi
%where $\mathcal{T}$ and $\mathcal{W}$ follow the relation $\mathcal{T}= \{ \bm s = \bm C \bm \theta + \bm d ~|~ \bm \theta \in \mathcal{W} \}$.
The proof of \eqref{eq:T_W_equiv} is analogous to that of Proposition~\ref{prop:mut}, and will not be repeated here.

{\it Step 3:} \
From the equivalent representation \eqref{eq:T_W_equiv}, we further deduce the following results:
i) $\mathcal{W}_e, \mathcal{W}' \in {\sf MVES}(\mathcal{B}(\sqrt{r^2 -1/N}))$ for $r= 1/\sqrt{N-1}$,
which is due to Step 1 and \eqref{eq:proof_suff_t0};
% where
%$\mu = \sqrt{ r^2 - 1/N } |_{r= 1/\sqrt{N-1}} = 1/\sqrt{(N-1)N}$;
ii) $\mathcal{W}_e \cap \mathcal{B}(\sqrt{r^2 -1/N}) \subseteq \mathcal{W}'$ for all $r > 1/\sqrt{N-1}$,
which is due to the underlying assumption that $\mathcal{T}' \in {\sf MVES}(\mathcal{R}(r))$ for $1/\sqrt{N-1} < r \leq 1$.
Consider the following lemma:
\begin{Lemma} \label{Property:TandTprime}
Suppose that $\setW, \setW'\in\MVES(\setB(r))$, where $\setB(r)$ is defined in \eqref{eq:Br_ball}.
Also, suppose that $\setR = \setW \cap
\setB(\bar{r})\subseteq\setW'$ for some $\bar{r}>r>0$. Then we have $\setW=\setW'$.
\end{Lemma}
The proof of Lemma~\ref{Property:TandTprime} is relegated to Appendix~\ref{proofsec:Property:TandTprime}.
By Lemma~\ref{Property:TandTprime}, we obtain $\mathcal{W}_e = \mathcal{W}'$, and consequently, $\mathcal{T}_e = \mathcal{T}'$.

\subsection{Proof of Proposition~\ref{prop:N2}}

{
Assume $N=2$, and
let ${\rm conv}\{ \bm b_1, \bm b_2 \}$ be an MVES of $\mathcal{S}_L$,
where $\bm b_1, \bm b_2 \in {\rm aff}\{ \bm e_1, \bm e_2 \} \subseteq \mathbb{R}^2$.
Using the simple fact ${\rm aff}\{ \bm e_1, \bm e_2 \} = \{ \bm s \in \mathbb{R}^2 ~|~ s_1 + s_2 = 1 \}$,
we can write
\[ \bm b_1 = \begin{bmatrix} \beta_1 \\ 1- \beta_1 \end{bmatrix},
\bm b_2 = \begin{bmatrix} \beta_2 \\ 1- \beta_2 \end{bmatrix},
\]
for some coefficients $\beta_1, \beta_2 \in \mathbb{R}$.
By the same spirit, every abundance vector $\bm s_n$ (for $N=2$) can be written as
\[
\bm s_n = \begin{bmatrix} \alpha_n \\ 1- \alpha_n \end{bmatrix}, \quad n=1,\ldots,L,
\]
where $ 0 \leq \alpha_n \leq 1$.
From the above expressions, it is easy to show that the MVES enclosing property $\bm s_n \in {\rm conv}\{ \bm b_1, \bm b_2 \}$ is equivalent to
\begin{equation} \label{eq:N2_alpha_bnd}
\beta_2 \leq \alpha_n \leq \beta_1, \quad n=1,\ldots,L,
\end{equation}
where we assume $\beta_1 \geq \beta_2$ w.l.o.g.
Moreover, from the simplex volume formula in \eqref{eq:vol_formula},
the volume of ${\rm conv}\{ \bm b_1, \bm b_2 \}$ is
\begin{equation} \label{eq:N2_volb1b2}
{\rm vol}( {\rm conv}\{ \bm b_1, \bm b_2 \})= \beta_1 - \beta_2.
\end{equation}
From \eqref{eq:N2_alpha_bnd}-\eqref{eq:N2_volb1b2},
it is immediate that ${\rm conv}\{ \bm b_1, \bm b_2 \}$ is a minimum volume simplex enclosing $\mathcal{S}_L$ if and only if
\begin{equation} \label{eq:N2_beta_cond}
\beta_2 = \min_{n=1,\ldots,L} \alpha_n, \quad \beta_1 = \max_{n=1,\ldots,L} \alpha_n.
\end{equation}
Now, consider perfect identifiability $\{ \bm b_1, \bm b_2 \} = \{ \bm e_1, \bm e_2 \}$,
which is equivalent to $\beta_1 = 1$, $\beta_2= 0$.
Putting the above conditions into \eqref{eq:N2_beta_cond},
we see that perfect identifiability is achieved if and only if the pure-pixel assumption holds; i.e.,
there exist two pixels, indexed by $n_1$ and $n_2$, such that $\bm s_{n_1} = \bm e_1$ and
$\bm s_{n_2} = \bm e_2$ (or $\alpha_{n_1} = 1$, $\alpha_{n_2} = 0$), resp.
}

{
\subsection{Proof of Theorem~\ref{thm:pure2}}
\label{sec:proof:thm:pure2}

Let
\begin{equation} \label{eq:pij}
\bm p_{ij} = \alpha \bm e_i + (1 - \alpha) \bm e_j,
\end{equation}
for $i,j \in \{ 1,\ldots,N \}, i \neq j$, and recall $\alpha= \min_{i \neq j} \alpha_{ij}$.
It can be verified that each $\bm p_{ij}$ is a convex combination of $\bm s_{n(i,j)}$ and $\bm s_{n(j,i)}$ in \eqref{eq:s_pure2}.
Thus, every $\bm p_{ij}$ satisfies $\bm p_{ij} \in {\rm conv} \mathcal{S}_L$.
For notational convenience,
let
\[
\mathcal{P}  = \{ \bm p_{ij} \}_{ i, j \in \{ 1,\ldots, N \}, ~ i \neq j }
\]
denote the set that collects all the $\bm p_{ij}$'s.
By the result $\bm p_{ij} \in {\rm conv} \mathcal{S}_L$,
we have
${\rm conv} \mathcal{P} \subseteq {\rm conv} \mathcal{S}_L$,
and consequently,
\[ \mathcal{R}(r) \subseteq {\rm conv} \mathcal{S}_L \Longleftarrow \mathcal{R}(r) \subseteq {\rm conv} \mathcal{P}.
\]
Applying the above implication to $\gamma$ in \eqref{eq:gamma_def} yields
\begin{equation} \label{eq:proof:pure2_1}
\gamma \geq \sup\{ r { \leq 1} ~|~ \mathcal{R}(r) \subseteq {\rm conv} \mathcal{P} \}
\end{equation}

Eq.~\eqref{eq:proof:pure2_1} has an explicit expression.
To show it, let us first consider the following lemma.

\begin{Lemma} \label{lem:PU}
For any $\alpha \in ( 0.5, 1]$, ${\rm conv} \mathcal{P}$ is equivalent to
\begin{equation} \label{eq:lem:PU}
{\rm conv} \mathcal{P} = \{ \bm s \in \mathcal{T}_e ~|~ s_i \leq \alpha, i=1,\ldots,N \}.
\end{equation}
\end{Lemma}
The proof of Lemma~\ref{lem:PU} is relegated to Appendix~\ref{proofsec:lem:PU}.
By using Lemma~\ref{lem:PU}, and observing the expressions of $\mathcal{R}(r)$ in \eqref{eq:gamma_def} and ${\rm conv} \mathcal{P}$ in \eqref{eq:lem:PU},
we see the following equivalence
\begin{align}
\mathcal{R}(r) \subseteq {\rm conv} \mathcal{P}
    & \Longleftrightarrow \max_{i=1,\ldots,N} s_i \leq \alpha \text{~for all $\bm s \in \mathcal{R}(r)$} \nonumber \\
    & \Longleftrightarrow \sup_{ \bm s \in \mathcal{R}(r) } \max_{i=1,\ldots,N} s_i \leq \alpha,  \label{eq:proof:pure2_2}
\end{align}
for $\frac{1}{\sqrt{N}} \leq r \leq 1$ (note that $\mathcal{R}(r) = \emptyset$ for $r  <  \frac{1}{\sqrt{N}}$).
Next, we solve the maximization problem in \eqref{eq:proof:pure2_2}.
The result is summarized in the following lemma.

\begin{Lemma} \label{lem:alpha_r}
Let
\begin{equation*}
\alpha^\star(r) = \sup_{ \bm s \in \mathcal{R}(r) } \max_{i=1,\ldots,N} s_i,
\end{equation*}
where $N \geq 2$ and $\frac{1}{\sqrt{N}} \leq r \leq 1$.
The optimal value $\alpha^\star(r)$ has a closed-form expression
\[ \alpha^\star(r) = \frac{1 + \sqrt{(N-1)(N r^2 - 1)}}{N}. \]
\end{Lemma}
The proof of Lemma~\ref{lem:alpha_r} is shown in Appendix~\ref{proofsece:lem:alpha_r}.
Now, by applying Lemma~\ref{lem:alpha_r} and \eqref{eq:proof:pure2_2} to \eqref{eq:proof:pure2_1}, we get
\begin{equation} \label{eq:proof:pure2_3}
\gamma \geq \sup\{  r \in [ 1/\sqrt{N}, 1 ] ~|~ \alpha^\star(r) \leq \alpha \}.
\end{equation}
By noting that $\alpha^\star(r)$ is an increasing function of $r \in [ 1/\sqrt{N}, 1 ]$,
we see that if there exists an $r \in [ 1/\sqrt{N}, 1 ]$ such that $\alpha^\star(r) = \alpha$,
then that $r$ attains the supremum in \eqref{eq:proof:pure2_3}.
It can be verified that the solution to $\alpha^\star(r) = \alpha$ is
\[ r = \sqrt{ \frac{1}{N} \left[   \frac{(N \alpha -1)^2}{N-1} + 1 \right]}, \]
and the above $r$ satisfies $r \in [ 1/\sqrt{N}, 1 ]$ for $0.5 < \alpha \leq 1$, $N \geq 2$.
Putting the above solution into \eqref{eq:proof:pure2_3}, we obtain the desired result in Theorem~\ref{thm:pure2}.
}

\section{Numerical Experiments}
\label{sec:sim}

In this section, we provide numerical simulation results that aim to verify the theoretical MVES identifiability results proven in the previous section.
The signals are generated by the following way.
The observed data set $\{ \bm x_1,\ldots,\bm x_L \}$ follows the basic model in \eqref{eq:basic_model}.
The endmember signature vectors $\bm a_1,\ldots,\bm a_N$ are selected from the U.S. geological survey (USGS) library~\cite{USGS2007}, and the number of spectral bands is $M= 224$.
The generation of the abundance vectors is similar to that in \cite{Chan2009}.
Specifically, we generate a large pool of random vectors following a Dirichlet distribution with parameter $\bm \mu = \frac{1}{N} \bm 1$,
and then select a number of $L$ such random vectors as the abundance set $\{ \bm s_1, \ldots, \bm s_L \}$.
During the selection, we do not choose vectors whose $2$-norm exceeds a given parameter $r$;
the reason of doing so is to allow us to control the pixel purity level of $\{ \bm s_1, \ldots, \bm s_L \}$ at or below $r$ in the simulations.
Note that if the number of pixels $L$ is large,
then one should expect that $r$ be close to the best pixel purity level $\rho$ and uniform pixel purity level $\gamma$.
In the simulations, we set $L=1,000$.

The simulation settings are as follows.
MVES is implemented by the alternating linear programming method in \cite{Chan2009}.
We measure its identification performance by using the root-mean-square (RMS) angle error
\[
\phi = \min_{ \bm \pi \in \Pi_N } \sqrt{ \frac{1}{N} \sum_{i=1}^N \left[ {\rm arccos}\left( \frac{ \bm a_i^T \hat{\bm a}_{\pi_i} }{ \| \bm a_i \| \cdot \| \hat{\bm a}_{\pi_i} \|}  \right) \right]^2  },
\]
where $\{ \hat{\bm a}_1,\ldots,\hat{\bm a}_N \}$ denotes the MVES estimate of the endmembers, and
%$\Pi_N = \{ \bm \pi \in \mathbb{R}^N ~|~ \pi_i \in \{ 1,\ldots,N \}, \pi_i \neq \pi_j, \forall i \neq j \}$ is the set of all permutations of $\{1,\ldots,N\}$.
$\Pi_N$ denotes the set of all permutations of $\{1,\ldots,N\}$.
A number of $50$ randomly generated realizations were run to evaluate the means and standard deviations of $\phi$.

\begin{figure}[htp!]
    \begin{center}
        \subfigure[$N=3$]{\resizebox{0.45\textwidth}{!}{
        \psfrag{r}[Bc][Bc]{\LARGE$r$}
        \psfrag{phi}[Bc][Bc]{\LARGE $\phi\textrm{ (degrees)}$}
        \psfrag{rho}[Bc][Bc]{\LARGE$~~~~~r=\frac{1}{\sqrt{2}}$}
        \psfrag{rm}[Bc][Bc]{\LARGE{$\frac{1}{\sqrt{3}}$}}
        \includegraphics{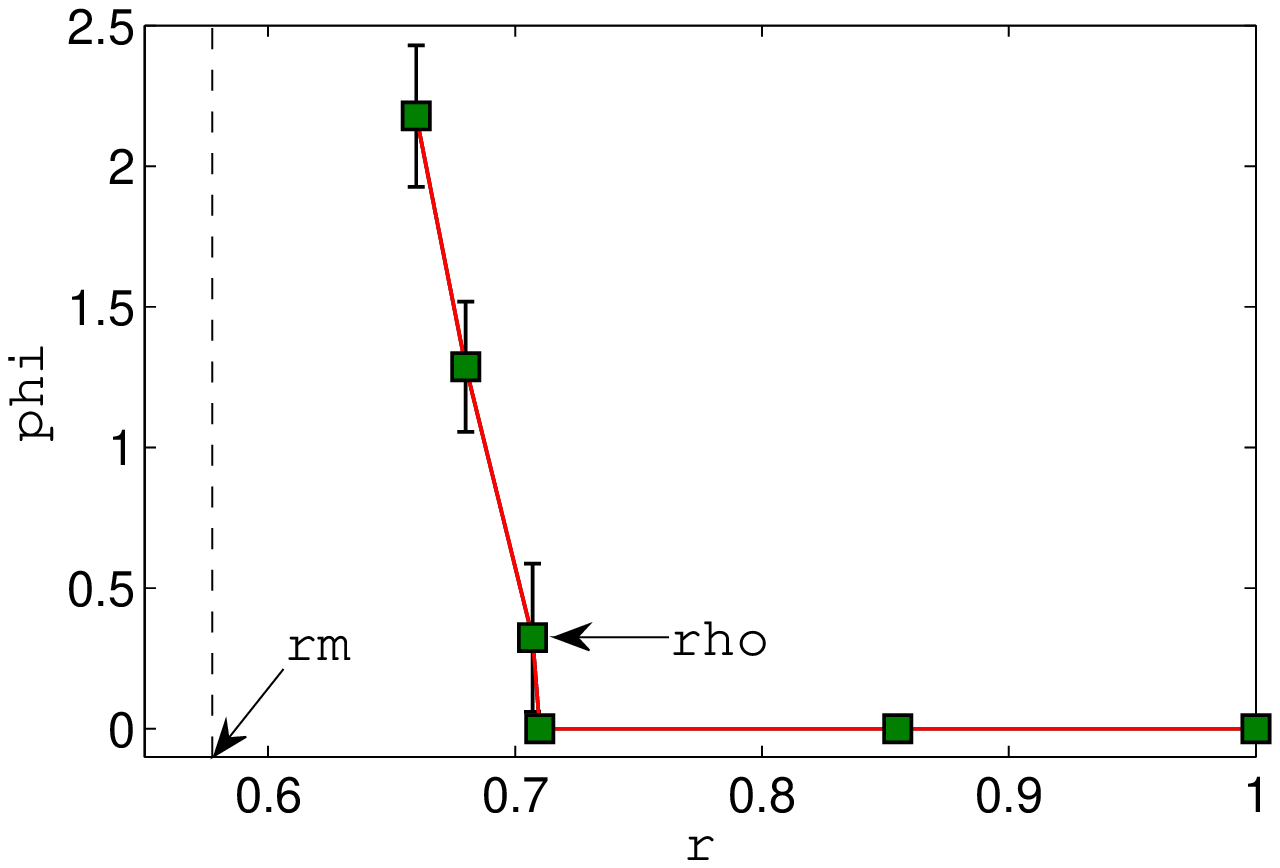}}}
        \subfigure[$N=4$]{\resizebox{0.45\textwidth}{!}{
        \psfrag{r}[Bc][Bc]{\LARGE$r$}
        \psfrag{phi}[Bc][Bc]{\LARGE $\phi\textrm{ (degrees)}$}
        \psfrag{rho}[Bc][Bc]{\LARGE$~~~~~r=\frac{1}{\sqrt{3}}$}
        \psfrag{rm}[Bc][Bc]{\LARGE{$\frac{1}{\sqrt{4}}$}}
        \includegraphics{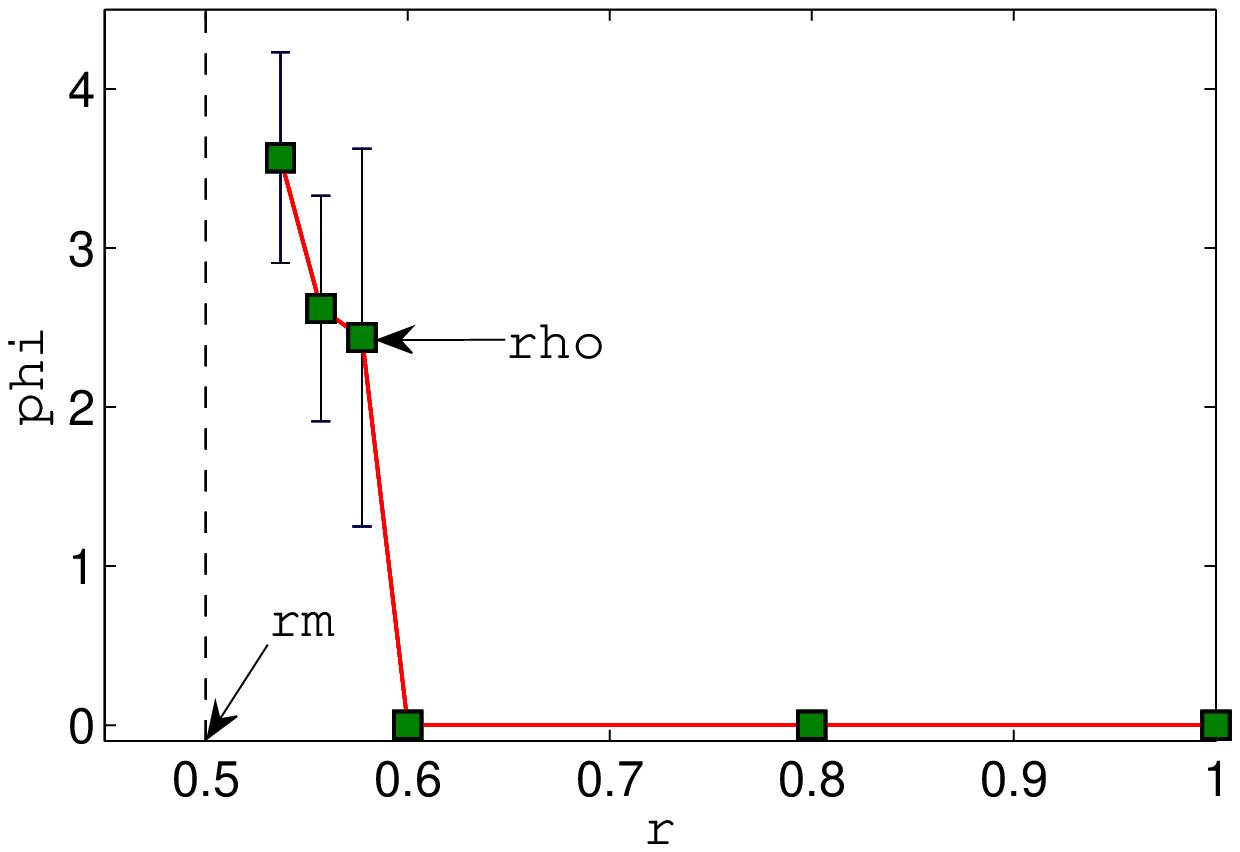}}}

        \subfigure[$N=5$]{\resizebox{0.45\textwidth}{!}{
        \psfrag{r}[Bc][Bc]{\LARGE$r$}
        \psfrag{phi}[Bc][Bc]{\LARGE $\phi\textrm{ (degrees)}$}
        \psfrag{rho}[Bc][Bc]{\LARGE$~~~~~r=\frac{1}{\sqrt{4}}$}
        \psfrag{rm}[Bc][Bc]{\LARGE{$\frac{1}{\sqrt{5}}$}}
        \includegraphics{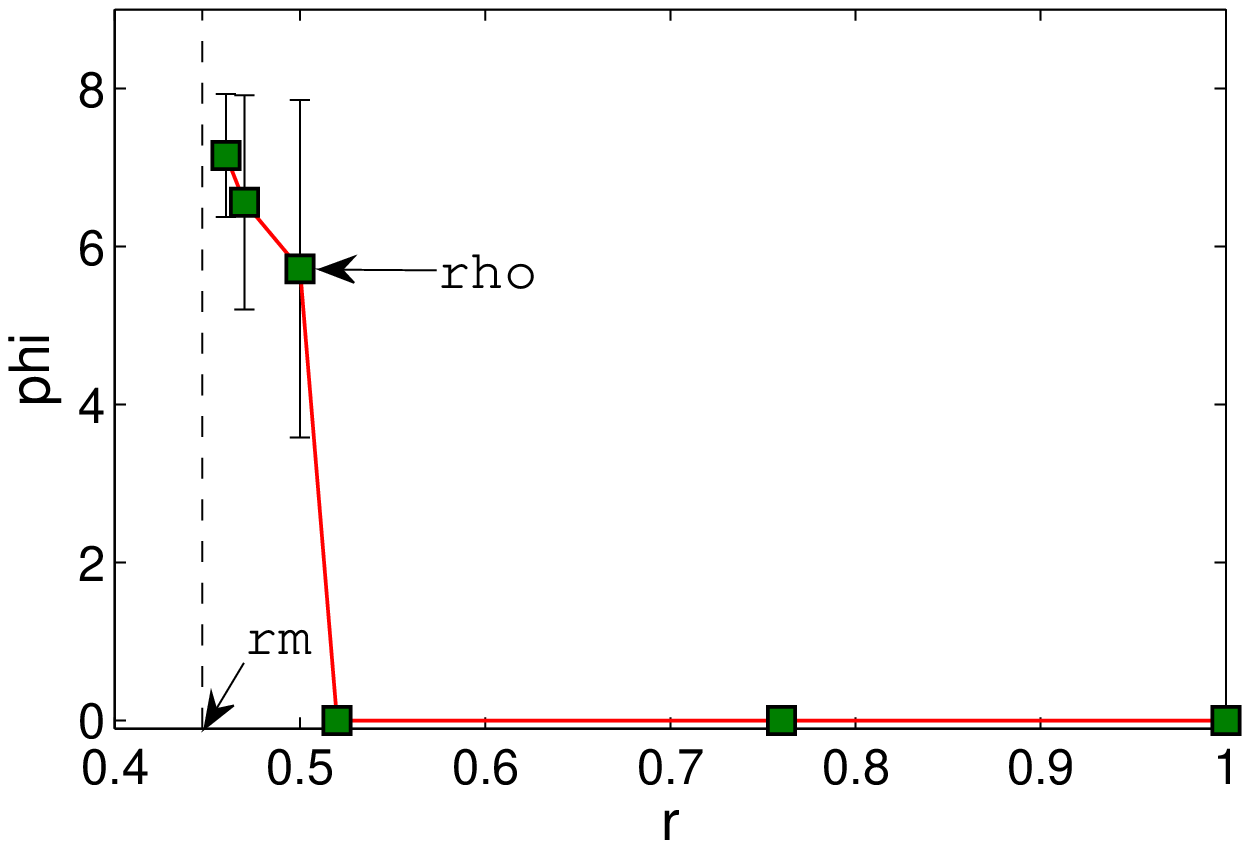}}}
        \subfigure[$N=6$]{\resizebox{0.45\textwidth}{!}{
        \psfrag{r}[Bc][Bc]{\LARGE$r$}
        \psfrag{phi}[Bc][Bc]{\LARGE $\phi\textrm{ (degrees)}$}
        \psfrag{rho}[Bc][Bc]{\LARGE$~~~~~r=\frac{1}{\sqrt{5}}$}
        \psfrag{rm}[Bc][Bc]{\LARGE{$\frac{1}{\sqrt{6}}$}}
        \includegraphics{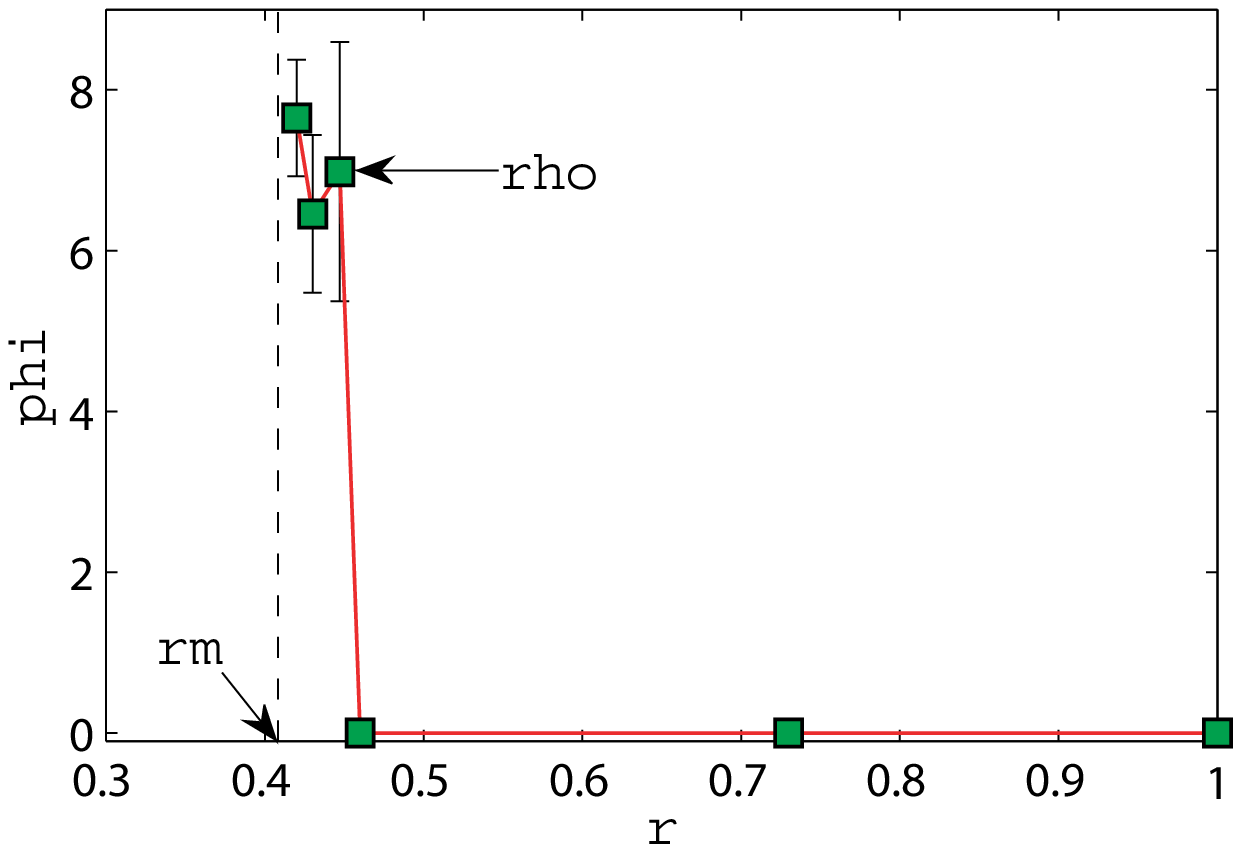}}}
    \end{center}
    \caption{MVES performance with respect to the numerically control pixel purity level $r$.}
    \label{fig:sim}
\end{figure}

The obtained RMS angle error results are shown in Figure~\ref{fig:sim}.
We see that zero RMS angle error, or equivalently, perfect identifiability, is attained when $r > 1/\sqrt{N-1}$ ---
which is a good match with the sufficient MVES identifiability result in Theorem~\ref{thm:suff}.
Also, we observe non-zero errors for $r \leq 1/\sqrt{N-1}$, which verifies the necessary MVES identifiability result in Theorem~\ref{thm:nec}.

Before closing this experiment section,
we should mention that previous papers, such as \cite{Chan2009,nascimento2012hyperspectral,plaza2012endmember,Arul2011,hendrix2012new,Lopes2010,agathos2014gpu}, have together provided a nice and rather complete coverage on MVES's performance under both synthetic and real-data experiments.
Hence, readers are referred to such papers for more experimental results.
The results reported therein also indicate that MVES-based algorithms are robust against lack of pure pixels.
The numerical (and also theoretical) results above further show the limit of robustness---$1/\sqrt{N-1}$ with the uniform pixel purity level.

\section{Conclusion}
\label{sec:con}

In this paper, a theoretical analysis for the identifiablility of MVES in blind HU was performed.
The results suggest that
under some mild assumptions which are considerably more relaxed than those for the pure-pixel case,
MVES exhibits robustness against lack of pure pixels.
Hence,
%this theoretical study provides an answer to the question
our study provides a theoretical explanation on
why numerical studies usually found that MVES can recover the endmembers accurately
%under intimately mixed data.
in the no pure-pixel case.
%{\blue \st{As future research, it would be interesting to investigate implications of the identifiability analysis on efficient algorithm designs, and extensions to even more challenging analysis scenarios such as noise sensitivity analysis.}}
%of the analysis such as noise sensitivity analysis.

\appendix

\ifplainver
    \section*{Appendix}
    \renewcommand{\thesubsection}{\Alph{subsection}}
\else
    \section{Appendix}
\fi

\subsection{Proof of  Lemma~\ref{lem:vol_trans}}
\label{proofsec:lem:vol_trans}

Let us first prove Lemma~\ref{lem:vol_trans}.(a).
The set $\mathcal{T}_G$ can be explicitly represented by
\begin{equation*}
\mathcal{T}_G = {\rm conv}\{ \bm g_1, \ldots, \bm g_N \},
\end{equation*}
where $\bm g_i \in \mathbb{R}^N$ for all $i$.
Also, by letting $\bm h_i = \bm A \bm g_i$ for all $i$, one can easily show that
\begin{equation*}
f(\mathcal{T}_G) = {\rm conv}\{ \bm h_1, \ldots, \bm h_N \}.
\end{equation*}
Since $\mathcal{T}_G \subset {\rm aff} \{ \bm e_1, \ldots, \bm e_N \}$, we have $\bm g_i \in {\rm aff} \{ \bm e_1, \ldots, \bm e_N \}$ for all $i$.
This means that each $\bm g_i$ satisfies $\bm 1^T \bm g_i = 1$, or equivalently, $g_{i,N} = 1 - \sum_{j=1}^{N-1} g_{i,j}$.
Using the above fact, we can write
\begin{equation*}
\bm g_i = \bm C {\bm \theta}_i + \bm e_N,
\end{equation*}
where ${\bm \theta}_i = [ \bm g_i ]_{1:(N-1)}$, and
\begin{equation*}
\bm C = %\begin{bmatrix}
%1      & 0 & \hdots & 0 \\
%0      & 1 &        & \vdots \\
%\vdots &   & \ddots &  0      \\
%0      & \hdots  & 0      & 1 \\
%-1     & -1      & \hdots & -1
%\end{bmatrix}.
\begin{bmatrix}
\bm I \\ - \bm 1^T
\end{bmatrix} \in \mathbb{R}^{N \times (N-1)}.
\end{equation*}
Let $\bar{\bm G} = [~ \bm g_1 - \bm g_N, \ldots, \bm g_{N-1} - \bm g_N ~]$. We get
\begin{equation*}
\bar{\bm G} = \bm C \bar{\bm \Theta},
\end{equation*}
where $\bar{\bm \Theta}= [~ {\bm \theta}_1 - {\bm \theta}_N, \ldots, {\bm \theta}_{N-1} - {\bm \theta}_N ~] \in \mathbb{R}^{(N-1) \times (N-1)}$.
We therefore obtain
\begin{subequations} \label{eq:detGG}
\begin{align}
%{\rm vol}(\mathcal{T}_G) = \frac{1}{(N-1)!} \sqrt{ \det( \bar{\bm G}^T \bar{\bm G} ) } =
\det( \bar{\bm G}^T \bar{\bm G} ) & = \det( \bar{\bm \Theta}^T \bm C^T \bm C \bar{\bm \Theta} ) \\
& = \det(\bar{\bm \Theta}) \det( \bm C^T \bm C ) \det( \bar{\bm \Theta} ) \label{eq:detGG-t1} \\
& = N \cdot | \det(\bar{\bm \Theta}) |^2,  \label{eq:detGG-t2}
\end{align}
\end{subequations}
where \eqref{eq:detGG-t1} is due to $\det( \bm A \bm B ) = \det(\bm A) \det(\bm B)$ for square $\bm A, \bm B$,
and \eqref{eq:detGG-t2} is due to the following result
\[
\det( \bm C^T \bm C ) = \det( \bm I + \bm 1 \bm 1^T ) = N
\]
(note that the matrix result $\det( \bm I + \bm q \bm q^T ) = \| \bm q \|^2 + 1$ has been used).
%$\det( \bm C^T \bm C ) = N$ (which can be easily proven).
Likewise, by letting $\bar{\bm H} = [~ \bm h_1 - \bm h_N, \ldots, \bm h_{N-1} - \bm h_N ~]$, we have
\begin{equation*}
\bar{\bm H} = \bm A \bar{\bm G} = \bm A \bm C \bar{\bm \Theta} = \bar{\bm A} \bar{\bm \Theta},
\end{equation*}
and
\begin{equation} \label{eq:detHH}
\det( \bar{\bm H}^T \bar{\bm H} ) = \det( \bar{\bm A}^T \bar{\bm A} ) \cdot | \det(\bar{\bm \Theta}) |^2.
\end{equation}
Now, by \eqref{eq:vol_formula}, \eqref{eq:detGG} and \eqref{eq:detHH}, Eq.~\eqref{eq:vol1} is obtained.
Also,
%Eq.~\eqref{eq:aff1}
the property $f(\mathcal{T}_G) \subset {\rm aff}\{ \bm a_1, \ldots, \bm a_N \}$
can be easily proven by the fact that $\bm H = \bm A \bm G$ and $\bm 1^T \bm g_i = 1$ for all $i$.

Next, we prove Lemma \ref{lem:vol_trans}.(b). The set $\mathcal{T}_H$ can be written as
\begin{equation*}
\mathcal{T}_H = {\rm conv}\{ \bm h_1, \ldots, \bm h_N \},
\end{equation*}
where $\bm h_i \in \mathbb{R}^M$ for all $i$.
Since $\mathcal{T}_H \subset {\rm aff} \{ \bm a_1, \ldots, \bm a_N \}$,
we have $\bm h_i \in {\rm aff}\{ \bm a_1, \ldots, \bm a_N \}$ for all $i$.
Hence, each $\bm h_i$ can be expressed as
%\begin{equation*}
$\bm h_i = \bm A \bm g_i$,
%\end{equation*}
where $\bm g_i \in \mathbb{R}^N$, $\bm 1^T \bm g_i =1$.
%Moreover, we have
This leads to
\begin{subequations}
\begin{align}
f^{-1}( \mathcal{T}_H ) & = \{ ~ \bm x ~|~ \bm A \bm x \in {\rm conv}\{ \bm h_1, \ldots, \bm h_N \} ~\}  \\
& = \{ ~ \bm x ~ | ~ \bm A \bm x = \bm H \bm \theta, ~ \bm \theta \geq \bm 0, \bm 1^T \bm \theta = 1 ~\} \\
& = \{ ~ \bm x ~ | ~ \bm A \bm x = \bm A \bm G \bm \theta, ~ \bm \theta \geq \bm 0, \bm 1^T \bm \theta = 1 ~\}  \\
& = \{ ~ \bm x ~ | ~ \bm x = \bm G \bm \theta, ~ \bm \theta \geq \bm 0, \bm 1^T \bm \theta = 1 ~\} \label{eq:ifTh_1} \\
& = {\rm conv}\{ \bm g_1, \ldots, \bm g_N \} \\
& \subset {\rm aff}\{ \bm e_1, \ldots, \bm e_N \}, \label{eq:ifTh_2}
\end{align}
\end{subequations}
where \eqref{eq:ifTh_1} is due to the full column rank condition of $\bm A$,
and \eqref{eq:ifTh_2} uses the structure $\bm 1^T \bm g_i =1$.
The rest of the proof is the same as that of Lemma \ref{lem:vol_trans}.(a).

\subsection{Proof of  Lemma~\ref{lem:Te_MVES}}
\label{proofsec:lem:Te_MVES}

Fix $r= 1/\sqrt{N-1}$.
From \eqref{eq:Cr_alt}, $\mathcal{C}(r)$ can be re-expressed as
\begin{equation} \label{eq:Cr_alt_alt}
\mathcal{C}(r) = \{ \bm s = \bm C \bm \theta + \bm d ~|~ \bm \theta \in \mathcal{B}(\mu) \},
\end{equation}
where $\mu = \sqrt{r^2 - 1/N } = 1/\sqrt{(N-1) N}$, and
\begin{equation} \label{eq:Br}
\mathcal{B}(r') = \{ \bm \theta \in \mathbb{R}^{N-1} ~|~ \| \bm \theta \| \leq r' \}
\end{equation}
is a ball on $\mathbb{R}^{N-1}$.
Also, recall from \eqref{eq:V_alt}-\eqref{eq:W_alt} that an MVES $\mathcal{V} \in {\sf MVES}(\mathcal{C}(r))$ can be written as
\begin{equation} \label{eq:V_alt_alt}
\mathcal{V} = \{ \bm s = \bm C \bm \theta + \bm d ~|~  \bm \theta \in \mathcal{W} \},
\end{equation}
where $\mathcal{W} = {\rm conv}\{ \bm w_1, \ldots, \bm w_N \} \subseteq \mathbb{R}^{N-1}$;
and that ${\rm vol}(\mathcal{V}) = {\rm vol}(\mathcal{W})$ (see \eqref{eq:vol_eq}).
From %\eqref{eq:Cr_alt_alt}, \eqref{eq:Br} and \eqref{eq:V_alt_alt},
the expressions above,
we can deduce the following result: $\mathcal{W}$ must be an MVES of $\mathcal{B}(\mu)$ if $\mathcal{V}$ is an MVES of $\mathcal{C}(r)$, and the converse is also true.

Next, we will use the following fact:
\begin{Fact} \label{fact:ball_mves} {\bf \cite[Theorem 3.2]{gerber1975}} The volume of an $(N-1)$-dimensional simplex $\mathcal{W}$ enclosing $B(r')$ in \eqref{eq:Br} satisfies
\begin{equation} \label{eq:fact_vol}
{\rm vol}( \mathcal{W} ) \geq \frac{1}{(N-1)!} N^{\frac{N}{2}} (N-1)^{\frac{1}{2}(N-1)} (r')^{N-1}
\end{equation}
with equality only for the regular simplex.
\end{Fact}
Using Fact~\ref{fact:ball_mves} and the result ${\rm vol}(\mathcal{V}) = {\rm vol}(\mathcal{W})$, we obtain
\begin{equation*}
{\rm vol}( \mathcal{V} ) = \frac{1}{(N-1)!} \sqrt{N},
\end{equation*}
where we should note that the right-hand side of the above equation is obtained by putting $r' = \mu = 1/\sqrt{(N-1) N}$ into \eqref{eq:fact_vol}.
On the other hand, consider $\mathcal{T}_e = {\rm conv}\{ \bm e_1, \ldots, \bm e_N \}$, which encloses $\mathcal{C}(r)$ (for $r= 1/\sqrt{N-1}$).
From the simplex volume formula \eqref{eq:vol_formula}, one can show that
\begin{equation*}
{\rm vol}( \mathcal{T}_e ) = \frac{1}{(N-1)!} \sqrt{N}.
\end{equation*}
Since $\mathcal{T}_e$ attains the same volume as $\mathcal{V}$, $\mathcal{T}_e$ is an MVES of $\mathcal{C}(r)$.

\subsection{Proof of Lemma~\ref{Property:TandTprime}}
\label{proofsec:Property:TandTprime}

The following lemma will be required:

\begin{Lemma} \label{lem:new}
Let $\mathcal{B}(r) = \{ \bm \theta \in \mathbb{R}^{N-1} ~|~  \| \bm \theta \| \leq r \}$,
where $r > 0$.
For any $\mathcal{W} \in {\sf MVES}(\mathcal{B}(r))$,
the boundaries of $\mathcal{B}(r)$ and  $\mathcal{W}$ have exactly $N$ intersecting points.
%Specifically, we have $\bd \mathcal{B}(r) \cap \bd \mathcal{W} = \{ \bm t_1, \ldots, \bm t_N \}$.
Also, by letting $\{ \bm t_1, \ldots, \bm t_N \}= \bd \mathcal{B}(r) \cap \bd \mathcal{W}$ be the set of those intersecting points,
we have the following properties:
\begin{itemize}
\item[(a)] The points $\bm t_1, \ldots, \bm t_N$ are affinely independent.
\item[(b)] The simplex $\mathcal{W}$ can be constructed from $\bm t_1, \ldots, \bm t_N$ via
\begin{equation*}
\mathcal{W} = \bigcap_{i=1}^N \left\{ \bm \theta \in \mathbb{R}^{N-1} ~|~ r^2  \geq \bm t_i^T \bm \theta \right\}.
\end{equation*}
\end{itemize}
%Any $\mathcal{W} \in {\sf MVES}(\mathcal{B}(r))$ has exactly $N$ intersecting points between $\bd \mathcal{B}(r)$ and $\bd \mathcal{W}$,
%and those points can be determined by \eqref{eq:t_pts} and \eqref{eq:polyh_simplex}.
\end{Lemma}
The proof of Lemma~\ref{lem:new} is given in Appendix~\ref{proofsece:lem:new}.
Let
\begin{align*}
\{\vct_1,\ldots,\vct_N\} & = \bd \mathcal{B}(r) \cap \bd \mathcal{W}, \\
\{\vct_1',\ldots,\vct_N'\} & = \bd \mathcal{B}(r) \cap \bd \mathcal{W}',
\end{align*}
which, by Lemma~\ref{lem:new}, always exist.
Since $\setB(r)\subset\setW$ and $\setB(r)\subset\setW'$,
%we can rewrite
the above two equations can be equivalently expressed as
%Let $\{\vct_1,\ldots,\vct_N\}$ (resp., $\{\vct_1',\ldots,\vct_N'\}$) be the $N$ intersecting points of $\bd\setB(r)$ and $\bd\setW$ (resp., $\bd\setW'$), and then from that $\setB(r)\subset\setW$ and $\setB(r)\subset\setW'$ ($\because$ $\setW, \setW'\in\MVES(\setB(r))$, we have
\begin{align}
\{\vct_1,\ldots,\vct_N\}&=
%\bd\setB(r)\cap\bd\setW=
\bd\setB(r)\setminus\inte\setW, \label{eq:proper3111}\\
\{\vct_1',\ldots,\vct_N'\}&=
%\bd\setB(r)\cap\bd\setW'=
\bd\setB(r)\setminus\inte\setW'.\label{eq:proper3111NEW}
\end{align}
Also, by Lemma~\ref{lem:new}.(b), we have $\setW=\setW'$ if $\{\vct_1,\ldots,\vct_N\}=\{\vct'_1,\ldots,\vct'_N\}$.
%To prove $\setW=\setW'$, it suffices to show that
%$\{\vct_1,\ldots,\vct_N\}=\{\vct'_1,\ldots,\vct'_N\}$.
% through the following steps.
%The latter is shown in the following steps.
In the following steps we focus on proving $\{\vct_1,\ldots,\vct_N\}=\{\vct'_1,\ldots,\vct'_N\}$.

\emph{Step $1$:}
We first prove
\begin{align} \label{eq:BDproper}
\bd\left(\setW\cap \setB(\bar{r})\right)\subseteq\bd\setW\cup
\bd\setB(\bar{r})
\end{align}
by contradiction. Suppose that \eqref{eq:BDproper} does not hold, namely, there exists an $\vcx\in\Rbb^{N-1}$ satisfying
\begin{align}
\vcx &\in\bd \left( \setW\cap
\setB(\bar{r}) \right) \textrm{, but}\label{eq:HK111}\\
\vcx &\notin\bd\setW\cup
\bd\setB(\bar{r}).\label{eq:HK222}
\end{align}
%However, the fact that
Now, since
$\setW\cap
\setB(\bar{r})$ is a closed set,
%together with \eqref{eq:HK111}, implies
\eqref{eq:HK111} implies
\begin{align}
\vcx \in\setW\cap
\setB(\bar{r}).\label{eq:HK333}
\end{align}
Equations \eqref{eq:HK222} and \eqref{eq:HK333} imply that $\vcx\in\inte\setW$ and that
$\vcx\in\inte\setB(\bar{r})$. Thus, we have
$\vcx\in\inte(\setW\cap\setB(\bar{r}))$ which contradicts \eqref{eq:HK111}. Hence,
\eqref{eq:BDproper} must hold.

\emph{Step $2$:}
We show that $\{\vct_1,\ldots,\vct_N\} = \bd\setB(r)\cap
\bd\setR$.
Let us first consider proving $\{\vct_1,\ldots,\vct_N\}\subseteq\bd\setB(r)\cap
\bd\setR$.
%We prove
%$\{\vct_1,\ldots,\vct_N\}\subseteq\bd\setB(r)\cap
%\bd\setR$ in this step.
We observe from $\setB(r)\subseteq\setB(\bar{r})$
%($\because$ $r<\bar{r}$)
and $\setB(r)\subseteq\setW$
%($\because$ $\setW\in\MVES(\setB(r))$)
that
\begin{align} \label{eq:BcriR}
\setB(r)\subseteq\setB(\bar{r})\cap\setW=\setR.
\end{align}
Subsequently, the following inequality chain can be derived:
%Then the desired result can be obtained through the following inequality chain:
%\begin{align} \label{eq:proper3222}
%\{\vct_1,\ldots,\vct_N\}
%=&\bd\setB(r)\setminus\inte\setW \textrm{~~~~~~~(by \eqref{eq:proper3111})} \nonumber\\
%\subseteq &\bd\setB(r)\setminus(\inte\setW\cap\inte\setB(\bar{r})) \nonumber\\
%=&\bd\setB(r)\setminus\inte\setR \textrm{~~~~~~~(by $\inte(\setW\cap\setB(\bar{r}))= \inte\setW\cap\inte\setB(\bar{r})$)} \nonumber\\
%=&\bd\setB(r)\cap \bd\setR. \textrm{~~~~~~~(by
%\eqref{eq:BcriR})}
%\end{align}
\begin{subequations} \label{eq:proper3222}
\begin{align}
\{\vct_1,\ldots,\vct_N\}
=&\bd\setB(r)\setminus\inte\setW  \label{eq:proper3222a} \\
%\textrm{~~~~~~~(by \eqref{eq:proper3111})} \\
\subseteq &\bd\setB(r)\setminus(\inte\setW\cap\inte\setB(\bar{r})) \\
=&\bd\setB(r)\setminus\inte\setR \label{eq:proper3222c} \\
%\textrm{~~~~~~~(by $\inte(\setW\cap\setB(\bar{r}))= \inte\setW\cap\inte\setB(\bar{r})$)} \nonumber\\
=&\bd\setB(r)\cap \bd\setR,   \label{eq:proper3222d}
%\textrm{~~~~~~~(by \eqref{eq:BcriR})}
\end{align}
\end{subequations}
where \eqref{eq:proper3222a} is by \eqref{eq:proper3111};
\eqref{eq:proper3222c} is by $\inte(\setW\cap\setB(\bar{r}))= \inte\setW\cap\inte\setB(\bar{r})$;
\eqref{eq:proper3222d} is by \eqref{eq:BcriR}.

%
%\emph{Step $3$:}
%Then prove
Moreover, we have
$\bd\setB(r)\cap\bd\setR\subseteq\{\vct_1,\ldots,\vct_N\}$, obtained from the following chain:
%\begin{align} \label{eq:proper3222111}
%\bd\setB(r)\cap \bd\setR
%=& \bd\setB(r)\cap \bd(\setW\cap \setB(\bar{r}))
%%\textrm{~~~~~~~(by the definition of $\setR$)}
%\nonumber\\
%\subseteq& \bd\setB(r)\cap \left(\bd\setW\cup \bd\setB(\bar{r})\right) \textrm{~~~~~~~(by \eqref{eq:BDproper})} \nonumber\\
%=&\left(\bd\setB(r) \cap \bd\setW\right)\cup \left(\bd\setB(r) \cap \bd\setB(\bar{r})\right) \nonumber\\
%=&\left(\bd\setB(r) \cap \bd\setW\right)\cup \emptyset \textrm{~~~~~~~(by $\bar{r}>r$)} \nonumber\\
%=&\bd\setB(r) \setminus \inte\setW \textrm{~~~~~~~(by $\bd\setB(r)\subseteq\setB(r)\subseteq\setW$)} \nonumber\\
%=&\{\vct_1,\ldots,\vct_N\}. \textrm{~~~~~~~(by
%\eqref{eq:proper3111})}
%\end{align}
\begin{subequations} \label{eq:proper3222111}
\begin{align}
\bd\setB(r)\cap \bd\setR
=& \bd\setB(r)\cap \bd(\setW\cap \setB(\bar{r})) \\
%\textrm{~~~~~~~(by the definition of $\setR$)}
\subseteq& \bd\setB(r)\cap \left(\bd\setW\cup \bd\setB(\bar{r})\right) \label{eq:proper3222111b} \\
%\textrm{~~~~~~~(by \eqref{eq:BDproper})} \nonumber\\
=&\left(\bd\setB(r) \cap \bd\setW\right)\cup \left(\bd\setB(r) \cap \bd\setB(\bar{r})\right) \\
=&\left(\bd\setB(r) \cap \bd\setW\right)\cup \emptyset \label{eq:proper3222111d} \\
%\textrm{~~~~~~~(by $\bar{r}>r$)}
=&\bd\setB(r) \setminus \inte\setW \label{eq:proper3222111e} \\
%\textrm{~~~~~~~(by $\bd\setB(r)\subseteq\setB(r)\subseteq\setW$)} \nonumber\\
=&\{\vct_1,\ldots,\vct_N\},  \label{eq:proper3222111f}
%\textrm{~~~~~~~(by \eqref{eq:proper3111})}
\end{align}
\end{subequations}
where \eqref{eq:proper3222111b} is by \eqref{eq:BDproper};
\eqref{eq:proper3222111d} is by $\bar{r}>r$;
\eqref{eq:proper3222111e} is by $\bd\setB(r)\subseteq\setB(r)\subseteq\setW$;
\eqref{eq:proper3222111f} is by \eqref{eq:proper3111}.

\emph{Step $3$:} We prove $\{\vct_1,\ldots,\vct_N\}=\{\vct'_1,\ldots,\vct'_N\}$.
%Combining \eqref{eq:proper3222} and \eqref{eq:proper3222111} yields
In Step $2$, it is shown that
\begin{align} \label{eq:proper3333}
\{\vct_1,\ldots,\vct_N\}=\bd\setB(r)\cap \bd\setR.
\end{align}
By the fact that
$\vct'_i\in\setB(r)$ and by \eqref{eq:BcriR}, we have
\begin{align} \label{eq:proper3aaaa}
\vct'_i\in
%\setB(r)\subseteq
\setR.
\end{align}
Moreover, from the assumption that $\setR \subseteq \setW'$, we have $\bd\setW'\cap\inte\setR=\emptyset$. But from \eqref{eq:proper3111NEW}, we note that $\vct'_i\in\bd\setW'$. Thus we can conclude $\vct'_i\notin
\inte(\setR)$, which together with \eqref{eq:proper3aaaa} yields
\begin{align} \label{eq:GOOD}
\vct'_i\in \bd\setR.
\end{align}
Combining $\vct'_i\in \bd\setB(r)$ (cf. \eqref{eq:proper3111NEW}) with \eqref{eq:proper3333} and \eqref{eq:GOOD}, we obtain $\vct'_i\in
\{\vct_1,\ldots,\vct_N\}$.
Since Property (a) in Lemma~\ref{lem:new} restricts $\vct'_1,\ldots,\vct'_N$ to be affinely independent, the only possible choice of $\vct'_1,\ldots,\vct'_N$ is  $\{\vct'_1,\ldots,\vct'_N\} = \{\vct_1,\ldots,\vct_N\}$.
%But we have $\vct'_i\neq \vct'_j$ for all $1\leq i<j \leq N$, and thus $\{\vct_1,\ldots,\vct_N\}=\{\vct'_1,\ldots,\vct'_N\}$.
Lemma \ref{Property:TandTprime} is therefore proven.

\subsection{Proof of Lemma~\ref{lem:new}}
\label{proofsece:lem:new}

The proof of Lemma~\ref{lem:new} requires several convex analysis results.
To start with, consider the following results:

\begin{Fact} \label{fact:poly_simplex}
Let $\mathcal{W} = {\rm conv}\{ \bm w_1, \ldots, \bm w_N \} \subset \mathbb{R}^{N-1}$ denote an $(N-1)$-dimensional simplex.
Also, let
\ifconfver
    \begin{equation} \label{eq:polyh}
    \begin{aligned}
    \mathcal{P}( \bm g, \bm H) & = \{ \bm \theta \in \mathbb{R}^{N-1} ~|~ \bm H^T \bm \theta + \bm g \geq \bm 0, \\
    &
    \quad\quad
    %\quad\quad\quad\quad\quad\quad
    - (\bm H \bm 1)^T \bm \theta + ( 1 - \bm 1^T \bm g ) \geq 0 \}
    \end{aligned}
    \end{equation}
\else
    \begin{align} \label{eq:polyh}
    \mathcal{P}( \bm g, \bm H) & = \{ \bm \theta \in \mathbb{R}^{N-1} ~|~ \bm H^T \bm \theta + \bm g \geq \bm 0, ~
    - (\bm H \bm 1)^T \bm \theta + ( 1 - \bm 1^T \bm g ) \geq 0 \}
    \end{align}
\fi
denote a polyhedron, where $(\bm g, \bm H) \in \mathbb{R}^{N-1} \times \mathbb{R}^{(N-1) \times (N-1)}$ is given.
\begin{itemize}
\item[(a)] Any $\mathcal{W}$ can be equivalently represented by $\mathcal{P}( \bm g, \bm H)$ via setting
\begin{align} \label{eq:polyh_simplex}
\bm H = \bar{\bm W}^{-T}, \quad \bm g = - \bar{\bm W}^{-T} \bm w_N,
\end{align}
where $\bar{\bm W} = [~ \bm w_1 - \bm w_N, \ldots, \bm w_{N-1} - \bm w_N ~]$.
\item[(b)] Suppose that $\bm H$ has full rank. Under the above restriction, the set $\mathcal{P}( \bm g, \bm H)$ for any $(\bm g, \bm H)$ can be equivalently represented by $\mathcal{W}$, whose vertices $\bm w_1, \ldots, \bm w_N$ can be determined by solving the inverse of \eqref{eq:polyh_simplex}.
    Also, the corresponding volume is
    \begin{equation} \label{eq:volP}
    {\rm vol}(\mathcal{P}(\bm g, \bm H)) = \frac{1}{(N-1)!} | \det(\bm H ) |^{-1}.
    \end{equation}
\end{itemize}
\end{Fact}

The proof of Fact~\ref{fact:poly_simplex} has been shown in the literature~\cite{boyd2004convex,Chan2009}.
Also, \eqref{eq:volP} is determined by the simplex volume formula \eqref{eq:vol_formula} and the relation in \eqref{eq:polyh_simplex}.
From Fact~\ref{fact:poly_simplex}, we derive several convex analysis properties for proving Lemma~\ref{lem:new}.

\begin{Fact} \label{fact:ball_simplex}
Let $\mathcal{W}$ be an $(N-1)$-dimensional simplex on $\mathbb{R}^{N-1}$, and consider the polyhedral representation of $\mathcal{W}$ in \eqref{eq:polyh}-\eqref{eq:polyh_simplex}.
Also, recall the definition $\mathcal{B}(r) = \{ \bm \theta \in \mathbb{R}^{N-1} ~|~ \| \bm \theta \| \leq r \}$.
\begin{itemize}
\item[(a)] If $\mathcal{B}(r) \subseteq \mathcal{W}$, then the following equations hold
\begin{subequations} \label{eq:ball_simplex1}
\begin{align}
- r \| \bm h_i \| + g_i & \geq 0, \quad i=1,\ldots,N-1, \label{eq:ball_simplex1a}  \\
- r \| \bm H \bm 1 \| + (1 - \bm 1^T \bm g ) & \geq 0, \label{eq:ball_simplex1b}
\end{align}
\end{subequations}
where $\bm h_i$ and $g_i$ denote the $i$th column of $\bm H$ and $i$th element of $\bm g$, resp.
Conversely, if \eqref{eq:ball_simplex1} holds, then $\mathcal{B}(r) \subseteq \mathcal{W}$.
\item[(b)] Suppose $\mathcal{B}(r) \subseteq \mathcal{W}$.
The boundaries of $\mathcal{B}(r)$ and  $\mathcal{W}$ have at most $N$ intersecting points.
Specifically, we have $\bd \mathcal{B}(r) \cap \bd \mathcal{W} \subseteq \{ \bm t_1, \ldots, \bm t_N \}$ where
%If there exists a point $\bm \theta$ that lies on the boundaries of both $\mathcal{B}(r)$ and  $\mathcal{W}$, or simply $\bm \theta \in \bd \mathcal{B}(r) \cap \bd \mathcal{W}$,
%    then $\bm \theta$ must be taken from $\{ \bm t_1, \ldots, \bm t_N \}$ where
    %\begin{align*}
%    - \frac{r}{ \| \bm h_1 \| } \bm h_i & \triangleq \bm t_1, \\
%    & \vdots \\
%    - \frac{r}{ \| \bm h_{N-1} \| } \bm h_{N-1} & \triangleq \bm t_{N-1}, \\
%    - \frac{r}{ \| \bm H \bm 1 \| } \bm H \bm 1 & \triangleq \bm t_N
%    \end{align*}
    \begin{subequations} \label{eq:t_pts}
    \begin{align}
    \bm t_i & = - \frac{r}{ \| \bm h_i \| } \bm h_i, \quad i=1,\ldots,N-1, \\
    \bm t_N & = \frac{r}{ \| \bm H \bm 1 \| } \bm H \bm 1.
    \end{align}
    \end{subequations}
    Also, if $\bm t_i \in \bd \mathcal{B}(r) \cap \bd \mathcal{W}$, then
    \begin{equation} \label{eq:ball_simplex2}
    \left\{ \begin{array}{ll}
    - r \| \bm h_i \| + g_i = 0, & \text{$i \in \{1,\ldots,N-1 \}$}, \\
    - r \| \bm H \bm 1 \| + (1 - \bm 1^T \bm g ) = 0, & \text{$i= N$};
    \end{array} \right.
    \end{equation}
    otherwise
    \begin{equation} \label{eq:ball_simplex3}
    \left\{ \begin{array}{ll}
    - r \| \bm h_i \| + g_i > 0, & \text{$i \in \{1,\ldots,N-1 \}$}, \\
    - r \| \bm H \bm 1 \| + (1 - \bm 1^T \bm g ) > 0, & \text{$i= N$}.
    \end{array} \right.
    \end{equation}
\end{itemize}
\end{Fact}

{\it Proof of Fact~\ref{fact:ball_simplex}:} \
%To prove Fact~\ref{fact:ball_simplex}.(a), note from \eqref{fact:poly_simplex} that the condition $\mathcal{B}(r) \subseteq \mathcal{W}$ can be equivalently expressed as
%\begin{subequations} \label{eq:pf_ball_simplex_1}
%\begin{align}
%\inf_{\| \bm \theta \| \leq r} \bm h_i^T \bm \theta + g_i & \geq 0, \quad i=1,\ldots,N-1,  \label{eq:pf_ball_simplex_1a} \\
%- \sup_{\| \bm \theta \| \leq r} (\bm H \bm 1)^T \bm \theta + ( 1 - \bm 1^T \bm g ) & \geq 0  \label{eq:pf_ball_simplex_1b}
%\end{align}
%\end{subequations}
%By the Cauchy-Schwartz inequality, we have $\bm h_i^T \bm \theta \geq - \| \bm h_i \| \| \bm \theta \| \geq - r \| \bm h_i \|$ for any $\| \bm \theta \| \leq r$, with the equality satisfied if and only if $\bm \theta = - ( r / \| \bm h_i \| ) \bm h_i$.
%Applying the above result to \eqref{eq:pf_ball_simplex_1a} yields \eqref{eq:ball_simplex1a}.
%Following the same proof as above, \eqref{eq:pf_ball_simplex_1b} is shown to be equivalent to \eqref{eq:ball_simplex1b}.
The proof of Fact~\ref{fact:ball_simplex}.(a) basically follows the development in \cite[pp.148-149]{boyd2004convex},
and is omitted here for conciseness.
To prove Fact~\ref{fact:ball_simplex}.(b), observe that a point $\tilde{\bm \theta} \in \bd \mathcal{B}(r) \cap \bd \mathcal{W}$ satisfies i) $\| \tilde{\bm \theta} \| = r$; and ii) either
\begin{equation} \label{eq:pf_ball_simplex_2a}
\bm h_i^T \tilde{\bm \theta} + g_i = 0,
\end{equation}
for some $i \in \{ 1,\ldots,N-1 \}$, or
\begin{equation} \label{eq:pf_ball_simplex_2b}
- (\bm H \bm 1)^T \tilde{\bm \theta} + ( 1 - \bm 1^T \bm g ) = 0.
\end{equation}
Suppose that $\tilde{\bm \theta}$ satisfies \eqref{eq:pf_ball_simplex_2a}.
Recall that the assumption $\mathcal{B}(r) \subseteq \mathcal{W}$ implies
\begin{align} \label{eq:pf_ball_simplex_3a}
\bm h_i^T \bm \theta + g_i & \geq 0, \quad \text{for all $\| \bm \theta \| \leq r$,}
\end{align}
and that the left-hand side of \eqref{eq:pf_ball_simplex_3a} attains its minimum if and only if  $\bm \theta = - ( r / \| \bm h_i \| ) \bm h_i = \bm t_i$.
Thus, if \eqref{eq:pf_ball_simplex_2a} is to be satisfied, then $\tilde{\bm \theta}$ must equal $\bm t_i$,
and subsequently \eqref{eq:pf_ball_simplex_2a} becomes
\begin{equation} \label{eq:pf_ball_simplex_4a}
- r \| \bm h_i \| + g_i = 0.
\end{equation}
Likewise, it is shown that if $\tilde{\bm \theta}$ satisfies \eqref{eq:pf_ball_simplex_2b}, then $\tilde{\bm \theta} = ( r/ \| \bm H \bm 1 \| ) \bm H \bm 1 = \bm t_N$ is the only choice and \eqref{eq:pf_ball_simplex_2b} becomes
\begin{equation} \label{eq:pf_ball_simplex_4b}
- r \| \bm H \bm 1 \| + (1 - \bm 1^T \bm g) = 0.
\end{equation}
We therefore complete the proof that $\tilde{\bm \theta} \in \bd \mathcal{B}(r) \cap \bd \mathcal{W}$ implies $\tilde{\bm \theta} \in \{ \bm t_1, \ldots, \bm t_N \}$.

We should also mention \eqref{eq:ball_simplex2}-\eqref{eq:ball_simplex3}.
From the proof above, it is clear that $\bm t_i \in \bd \mathcal{B}(r) \cap \bd \mathcal{W}$ holds if and only if \eqref{eq:pf_ball_simplex_4a} holds for $i=1,\ldots,N-1$, and \eqref{eq:pf_ball_simplex_4b} holds for $i= N$, respectively.
By considering \eqref{eq:ball_simplex1} as well, we obtain the conditions in \eqref{eq:ball_simplex2}-\eqref{eq:ball_simplex3}.
\hfill $\blacksquare$

\bigskip
We are now ready to prove Lemma~\ref{lem:new}.
Recall that $\mathcal{W} \in {\sf MVES}(\mathcal{B}(r))$ is assumed.
By Fact~\ref{fact:poly_simplex}.(a), we can write $\mathcal{W}= \mathcal{P}(\bm g, \bm H)$ for some $(\bm g, \bm H)$,
with $\bm H$ being of full rank.
Then, by Fact~\ref{fact:ball_simplex}.(b), we obtain $\bd \mathcal{B}(r) \cap \bd \mathcal{W} \subseteq \{ \bm t_1, \ldots, \bm t_N \}$.
%Suppose that $\bd \mathcal{B}(r) \cap \bd \mathcal{W}$ contains less than $N$ points.
We consider two cases.

{\it Case 1:} \
Suppose that $\bm t_i \notin \bd \mathcal{B}(r) \cap \bd \mathcal{W}$ for some $i \in \{ 1,\ldots,N-1 \}$.
For simplicity but
%without loss of generality,
{w.l.o.g.},
assume $i=1$.
By Fact~\ref{fact:ball_simplex}.(a)-(b),
we have
\begin{subequations} \label{eq:lem:new_case1}
\begin{align}
- r \| \bm h_1 \| + g_1 & > 0,  \\
- r \| \bm h_i \| + g_i & \geq 0, \quad i=2,\ldots,N-1, \\
- r \| \bm H \bm 1 \| + (1 - \bm 1^T \bm g ) & \geq 0.
\end{align}
\end{subequations}
Let us construct another polyhedron, denoted by $\mathcal{P}(\tilde{\bm g}, \tilde{\bm H} )$,
where the $2$-tuple $(\tilde{\bm g}, \tilde{\bm H}) \in \mathbb{R}^{N-1} \times \mathbb{R}^{(N-1) \times (N-1)}$ is chosen as
\begin{subequations} \label{eq:lem:new_case1_3}
\begin{align}
\tilde{g}_1 & = g_1 - N \epsilon,  \\
\tilde{g}_i & = g_i + \epsilon, \quad i=2,\ldots,N-1, \\
\tilde{\bm H} & = \left( \frac{r + \delta}{r} \right) \bm H, \label{eq:lem:new_case1_3c}
\end{align}
\end{subequations}
where
\begin{align}
\epsilon & = \frac{ - r \| \bm h_1 \| + g_1 }{2 N } > 0, \label{eq:lem:new_case1_1} \\
\delta   & = \frac{ \epsilon }{ \max\{ \| \bm h_1 \|, \ldots, \| \bm h_{N-1} \|, \| \bm H \bm 1 \| \} } > 0. \label{eq:lem:new_case1_2}
\end{align}
The polyhedron $\mathcal{P}(\tilde{\bm g}, \tilde{\bm H} )$ is also an $(N-1)$-dimensional simplex; this is shown by Fact~\ref{fact:poly_simplex}.(b) and the fact that the rank of $\tilde{\bm H}$ is the same as that of $\bm H$ (which is full).
Now, we claim that $\mathcal{B}(r) \subseteq \mathcal{P}(\tilde{\bm g}, \tilde{\bm H})$ and ${\rm vol}(\mathcal{P}(\tilde{\bm g}, \tilde{\bm H})) < {\rm vol}(\mathcal{P}({\bm g}, {\bm H})) = {\rm vol}(\mathcal{W})$.
For the first claim, one can verify from \eqref{eq:lem:new_case1}-\eqref{eq:lem:new_case1_3} that
\begin{align*}
- r \| \tilde{\bm h}_1 \| + \tilde{g}_1 & \geq (N-1) \epsilon \geq 0, \\
- r \| \tilde{\bm h}_i \| + \tilde{g}_i & \geq 0, \quad i=2,\ldots,N-1, \\
- r \| \tilde{\bm H} \bm 1 \| + (1 - \bm 1^T \tilde{\bm g}) & \geq \epsilon \geq 0,
\end{align*}
where $\tilde{\bm h}_i$ and $\tilde{g}_i$ denote the $i$th column of $\tilde{\bm H}$ and $i$th element of $\tilde{\bm g}$, resp.
The above equations, together with Fact~\ref{fact:ball_simplex}.(a), implies that $\mathcal{B}(r) \subseteq \mathcal{P}(\tilde{\bm g}, \tilde{\bm H})$.
The second claim follows from \eqref{eq:volP} in Fact~\ref{fact:poly_simplex}.(b) and \eqref{eq:lem:new_case1_3c}:
\begin{align}
{\rm vol}( \mathcal{P}(\tilde{\bm g}, \tilde{\bm H} ))
%& = \frac{1}{(N-1)!} | \det( \tilde{\bm H} ) |^{-1} \nonumber \\
& = \frac{1}{(N-1)!} \left( \frac{r}{r + \delta} \right)^{N-1}  | \det( {\bm H} ) |^{-1} \nonumber \\
& < \frac{1}{(N-1)!}  | \det( {\bm H} ) |^{-1} = {\rm vol}( \mathcal{W} ), \label{eq:lem:new_case1_4}
\end{align}
for $N \geq 2$ (note that $N=1$ is meaningless).
The above two claims contradicts the assumption that $\mathcal{W}$ is an MVES of $\mathcal{B}(r)$.

{\it Case 2:} \
Suppose that $\bm t_N \notin \bd \mathcal{B}(r) \cap \bd \mathcal{W}$.
The proof is similar to that of Case 1.
Very concisely, this case has $- r \| \bm H \bm 1 \| + (1 - \bm 1^T \bm g ) > 0$ and $- r \| \bm h_i \| + g_i \geq 0$ for all $i\in\{1,\ldots,N-1\}$.
By constructing a polyhedron $\mathcal{P}(\tilde{\bm g}, \tilde{\bm H})$ where
\begin{align*}
\tilde{\bm g} & = \bm g + \epsilon \bm 1, \quad
\tilde{\bm H}  = \left( \frac{r + \delta}{r} \right) \bm H, \\
\epsilon & = \frac{ - r \| \bm H \bm 1 \| + (1 - \bm 1^T \bm g ) }{2 N },
\end{align*}
and $\delta$ is the same as \eqref{eq:lem:new_case1_2},
we show that $\mathcal{B}(r) \subseteq \mathcal{P}(\tilde{\bm g}, \tilde{\bm H})$ and ${\rm vol}( \mathcal{P}(\tilde{\bm g}, \tilde{\bm H} )) < {\rm vol}( \mathcal{W} )$.
The above two claims contradict the MVES assumption with $\mathcal{W}$.

The above two cases imply that $\bd \mathcal{B}(r) \cap \bd \mathcal{W} = \{ \bm t_1, \ldots, \bm t_N \}$, the desired result.
In addition to this, Property (a) in Lemma~\ref{lem:new} is obvious since the expression of $\bm t_i$'s in \eqref{eq:t_pts}, as well as  \eqref{eq:polyh_simplex}, already suggest the affine independence of $\bm t_1, \ldots, \bm t_N$.
As for Property (b) in Lemma~\ref{lem:new}, note that \eqref{eq:ball_simplex2} are all satisfied.
It can be verified that by substituting \eqref{eq:t_pts} and \eqref{eq:ball_simplex2} into \eqref{eq:polyh}, $\mathcal{W}$ can be rewritten as $\mathcal{W}= \cap_{i=1}^N \left\{ \bm \theta \in \mathbb{R}^{N-1} ~|~ r^2  \geq \bm t_i^T \bm \theta \right\}$.

{
\subsection{Proof of Lemma~\ref{lem:PU}}
\label{proofsec:lem:PU}

For notational convenience, denote
\[ \mathcal{U}(\alpha) = \{ \bm s \in \mathcal{T}_e ~|~ s_i \leq \alpha, i=1,\ldots,N \}, \]
and recall that the aim is to prove ${\rm conv} \mathcal{P} = \mathcal{U}(\alpha)$.
The above identity is trivial for the case of $\alpha= 1$,
since we have ${\rm conv} \mathcal{P}=  \mathcal{T}_e \equiv \mathcal{U}(1)$ for $\alpha =1$.
Hence, we focus on $0.5 < \alpha < 1$.
The proof is split into three steps.

{\it Step 1:} We start with showing that $\bm s \in {\rm conv} \mathcal{P} \Longrightarrow \bm s \in \mathcal{U}(\alpha)$.
Note that any $\bm s \in {\rm conv} \mathcal{P}$ can be written as
\[ \bm s = \sum_{j \neq i} \theta_{ji} \bm p_{ij}, \]
for some $\{ \theta_{ji} \}$ satisfying $\sum_{j \neq i} \theta_{ji}= 1$ and $\theta_{ji} \geq 0$ for all $j,i$, $j \neq i$.
From the above equation and the expression of $\bm p_{ij}$ in \eqref{eq:pij},
one can verify that
%$\bm s \in {\rm conv}\{ \bm e_1, \ldots, \bm e_N \}$,
$\bm s \in \mathcal{T}_e$,
and that $s_k \leq \max_{j \neq i} [ \bm p_{ij} ]_k \leq \alpha$ for any $k$ (here $[ \bm p_{ij} ]_k$ denotes the $k$th element of $\bm p_{ij}$).
Thus, any $\bm s \in {\rm conv} \mathcal{P}$ also lies in $\mathcal{U}(\alpha)$.

%\medskip
%\noindent
{\it Step 2:} We turn our attention to proving $\bm s \in \mathcal{U}(\alpha) \Longrightarrow \bm s \in {\rm conv} \mathcal{P} $.
%To prove the implication $\bm s \in \mathcal{U}(\alpha) \Longrightarrow \bm s \in {\rm conv} \mathcal{P} $,
To proceed, suppose that $\bm s \in \mathcal{U}(\alpha)$,
%We consider $\alpha < 1$, since the case of $\alpha = 1$ is trivial (specifically, for $\alpha=1$, ${\rm conv} \mathcal{P}=  {\rm conv} \{ \bm e_1, \ldots, \bm e_N \} \equiv \mathcal{U}(1)$).
and assume $s_1 \geq s_2 \geq \ldots \geq s_N$
w.l.o.g.
%with loss of generality.
From a given $\bm s$,
%let us
choose an index $k$ by the following way
\begin{equation}  \label{eq:kk}
k = \max\{ i \in \{ 1,\ldots,N \} ~|~ s_i \geq \delta_i \},
\end{equation}
where $\delta_1 = 0$, and
\begin{equation} \label{eq:delta}
\delta_i = \frac{ 1 - \alpha - \sum_{j=i+1}^N s_j }{i-1},
\quad i=2,\ldots,N.
\end{equation}
From \eqref{eq:kk}-\eqref{eq:delta},
the following properties can be shown.
\begin{itemize}
\item[i)] It holds true that
\begin{equation} \label{eq:sord}
\begin{aligned}
s_1 & \geq \delta_k, \\
& \vdots  \\
s_k & \geq \delta_k, \\
s_{k+1} & < \delta_{k+1}, \\
& \vdots \\
s_N & < \delta_N.
\end{aligned}
\end{equation}
\item[ii)]
%$\sum_{j=k+1}^N s_j < 1 - \alpha$ for any $2 \leq k \leq N-1$.
{Suppose that $2 \leq k \leq N-1$, and $N \geq 3$. Then $\bm s$ satisfies $\sum_{j=k+1}^N s_j < 1 - \alpha$.}
\item[iii)] For any $\bm s \in \mathcal{U}(\alpha)$, the index $k$ must satisfy $k \geq 2$.
\item[iv)] $\alpha - \delta_k > 0$ for any $0.5 < \alpha \leq 1$.

\end{itemize}
The proofs of the above properties are as follows.
Property i) follows directly from the definition of $k$ and the ordering of $\bm s$.
{Property ii) is obtained by induction.
Observe that if $k \leq N-1$, the last equation of \eqref{eq:sord} reads
\begin{equation} \label{eq:sN_bnd}
s_N < \delta_N = \frac{1-\alpha}{N-1} \leq 1 - \alpha,
\end{equation}
and for $k= N-1$ the proof is complete (trivially).
For $k < N-1$,
we wish to show from \eqref{eq:sN_bnd} that $s_{N-1} + s_N < 1- \alpha$, and then recursively,
$\sum_{j=i}^N s_j < 1 - \alpha$ from $i=N-2$ to $i=k+1$.
To put this induction into context, suppose that
\begin{equation} \label{eq:s_psum_0}
\sum_{j=i+1}^N s_j < 1 - \alpha
\end{equation}
for $i \in \{ k+1,\ldots, N-1 \}$,
and note that \eqref{eq:s_psum_0} already holds for $i=N-1$ due to \eqref{eq:sN_bnd}.
The task is to prove $\sum_{j=i}^N s_j < 1 - \alpha$.
The proof is as follows:
\begin{subequations} \label{eq:s_sum_main}
\begin{align}
\sum_{j=i}^N s_j & < \delta_i + \sum_{j=i+1}^N s_j \label{eq:s_psum_a} \\
& = \frac{1-\alpha}{i-1} + \left( 1  - \frac{1}{i-1} \right) \sum_{j=i+1}^N s_j \label{eq:s_psum_b} \\
& < 1 - \alpha, \label{eq:s_psum_c}
\end{align}
\end{subequations}
where \eqref{eq:s_psum_a} is obtained by $s_i < \delta_i$ in Property i);
\eqref{eq:s_psum_b} by \eqref{eq:delta};
\eqref{eq:s_psum_c} by \eqref{eq:s_psum_0}, and $i-1 \geq k > 1$ for $k \geq 2$.
%It follows by induction that Property ii) holds.
%By applying \eqref{eq:s_sum_main} recursively up to $i=k+1$, Property ii) is obtained.
Hence, we conclude by induction that Property ii) holds.
}
To prove Property iii),
note that $\bm s$ satisfies $\bm 1^T \bm s = 1$.
Thus, $s_2$ can be written as
\begin{align*}
s_2 & = 1 - s_1 - \sum_{j=3}^N s_j
\end{align*}
Since every $\bm s \in \mathcal{U}(\alpha)$ satisfies $s_i \leq \alpha$ for any $i$, we get
\begin{align*}
s_2 & \geq 1 - \alpha - \sum_{j=3}^N s_j = \delta_2.
\end{align*}
The above condition implies that $k \geq 2$ must hold.
To prove Property iv),
observe the following inequalities
\begin{align*}
%\alpha - \delta_k & \geq  \alpha - \frac{1-\alpha}{k-1} \\
%& \geq \frac{2 \alpha - 1 }{k-1};
\alpha - \delta_k & \geq  \alpha - \frac{1-\alpha}{k-1} \geq \frac{2 \alpha - 1 }{k-1};
\end{align*}
here, the first inequality is done by applying \eqref{eq:delta},
 %and the property $\sum_{j=k+1}^N s_j \geq 0$,
and the second inequality by $k \geq 2$.
From the above equation, we see that $\alpha - \delta_k > 0$ for $\alpha > 0.5$.

With the above properties, we are ready to show that $\bm s \in \mathcal{U}(\alpha)$ lies in ${\rm conv} \mathcal{P}$.
First, for each $i \in \{ 1,\ldots, k \}$, we construct a vector
\[ \bar{\bm p}_i = \sum_{j \neq i} \theta_{ji} \bm p_{ij}, \]
where
\begin{align*}
\theta_{ji} & = \left\{ \begin{array}{ll}
c, & 1 \leq j \leq k, j \neq i \\
\displaystyle \frac{s_j}{1-\alpha}, & k+1 \leq j \leq N, { N \geq 3},
\end{array}
\right.  \\
c & = \frac{1}{k-1} \left( 1 - \frac{\sum_{j=k+1}^N s_j}{1-\alpha} \right) = \frac{\delta_k}{1-\alpha}.
\end{align*}
It can be verified that $\theta_{ji} \geq 0$, $\sum_{j\neq i} \theta_{ji} = 1$ (in particular, Property ii) is required to verify $c > 0$);
that is to say, every $\bar{\bm p}_i$ satisfies $\bar{\bm p}_i \in {\rm conv}\mathcal{P}$.
Moreover, from the above equations, $\bar{\bm p}_i$ is shown to take the structure
\begin{equation} \label{eq:bar_p_struc}
\bar{\bm p}_i = \begin{bmatrix}
( \alpha - \delta_k ) \bm e_i + \delta_k \bm 1 & \\
\bm s_{k+1:N}
\end{bmatrix},
\end{equation}
where $\bm s_{k+1:N} = [~ s_{k+1},\ldots, s_N ~]^T$.
Now, we claim that
\begin{equation} \label{eq:s_bar_p}
\bm s = \sum_{i=1}^k \beta_i \bar{\bm p}_i,
\end{equation}
where
\begin{equation} \label{eq:betas}
\beta_i = \frac{s_i - \delta_k}{\alpha - \delta_k}, ~ i=1,\ldots,k,
\end{equation}
and they satisfy
$\sum_{i=1}^k \beta_i = 1$, $\beta_i \geq 0$ for all $i$.
The above claim is verified as follows.
The property $\beta_i \geq 0$ directly follows from Properties i) and iv).
For the property $\sum_{i=1}^k \beta_i = 1$, observe that
\begin{align*}
\sum_{i=1}^k \beta_i & = \frac{ \sum_{i=1}^k s_i  - k \delta_k }{ \alpha - \delta_k } \\
& = \frac{ 1 - \sum_{j=k+1}^N s_j  - k \delta_k }{ \alpha - \delta_k } \\
& = \frac{ (k-1) \delta_k + \alpha - k \delta_k }{ \alpha - \delta_k } = 1,
\end{align*}
where the second equality is by $\bm 1^T \bm s = 1$,
and the third equality by \eqref{eq:delta}.
In addition, by substituting \eqref{eq:bar_p_struc} and \eqref{eq:betas} into the right-hand side of \eqref{eq:s_bar_p}, and by using $\bm 1^T \bm s = 1$, one can show that \eqref{eq:s_bar_p} is true.
Eq.~\eqref{eq:s_bar_p} and the associated properties with $\beta_i$ suggest that
$\bm s \in {\rm conv}\{ \bar{\bm p}_1,\ldots,\bar{\bm{p}}_k \}$.
This, together with the fact that $\bar{\bm p}_i \in {\rm conv}\mathcal{P}$,
implies $\bm s \in {\rm conv} \mathcal{P}$.

%\medskip
%\noindent
{\it Step 3:}
By combining the results in Step 1 and Step 2, we get $\bm s \in {\rm conv} \mathcal{P} \Longleftrightarrow \bm s \in \mathcal{U}(\alpha)$.
Lemma~\ref{lem:PU} is therefore proven.

\subsection{Proof of Lemma~\ref{lem:alpha_r}}
\label{proofsece:lem:alpha_r}

Recall $\mathcal{R}(r) = \{ \bm s \in \mathcal{T}_e ~|~ \| \bm s \| \leq r \}$,
and notice that $\mathcal{T}_e$ can be rewritten as
\[ \mathcal{T}_e= \{ \bm s \in \mathbb{R}^N ~|~ \bm s \geq \bm 0, \bm 1^T \bm s = 1 \}. \]
Let $\bm s \in \mathcal{R}(r)$, and assume $s_1 \geq s_2 \geq \ldots \geq s_N$ w.l.o.g.
From the above assumption, it is easy to verify that $s_1 \geq \frac{1}{N}$.
Also, by denoting $\bm s_{2:N} = [~ s_2,\ldots,s_N ~]^T$, we have
\begin{align}
r^2 & \geq \| \bm s \|^2
%\nonumber \\
%&
= s_1^2 + \| \bm s_{2:N} \|^2 \nonumber \\
& \geq s_1^2 + \frac{(1-s_1)^2}{N-1} \label{eq:lem:alpha_r_ineq1}
\end{align}
where the second inequality is owing to the norm inequality
$\sum_{i=1}^n |x_i| \leq \sqrt{n} \| \bm x \|$ for any $\bm x \in \mathbb{R}^n$,
and the fact that $\bm s \geq \bm 0$, $\bm 1^T \bm s = 1$.
Moreover, equality in \eqref{eq:lem:alpha_r_ineq1} holds if $\bm s$ takes the form
$\bm s = [~ s_1, \frac{1-s_1}{N-1} \bm 1^T ~]^T$ (which lies in $\mathcal{T}_e$).
Hence, $\alpha^\star(r)$ can be simplified to
\begin{subequations} \label{eq:lem:alpha_r_prob1}
\begin{align}
\alpha^\star(r) = \sup & ~ s_1 \\
{\rm s.t.} & ~ s_1^2 + \frac{(1-s_1)^2}{N-1} \leq r^2  \label{eq:lem:alpha_r_prob1b} \\
& ~ \frac{1}{N} \leq s_1 \leq 1. \label{eq:lem:alpha_r_prob1c}
\end{align}
\end{subequations}
By the quadratic formula, the constraint in \eqref{eq:lem:alpha_r_prob1b} can be reexpressed as
\begin{equation} \label{eq:lem:alpha_r_prob1b_eq1}
\left( s_1  - a \right)
\left( s_1  - b \right) \leq 0,
\end{equation}
where
\begin{align*}
a & = \frac{ 1 + \sqrt{(N-1)(Nr^2 -1)}}{N}, \\
b & = \frac{ 1 - \sqrt{(N-1)(Nr^2 -1)}}{N}.
\end{align*}
%and notice that for $ \frac{1}{\sqrt{N}} < r \leq 1$,
%we have $a \leq 1$ and $b \leq \frac{1}{N}$,
%and that for $r = \frac{1}{\sqrt{N}}$, we have $a= b = \frac{1}{N}$.
%Using the above conditions and \eqref{eq:lem:alpha_r_prob1c}, it is verified that \eqref{eq:lem:alpha_r_prob1b_eq1} can be reduced to $s_1 \leq a$.
%By substituting $s_1 \leq a$ back to \eqref{eq:lem:alpha_r_prob1b}, we get $\alpha^\star(r) = a$.
%The proof is therefore complete.
From \eqref{eq:lem:alpha_r_prob1c} and \eqref{eq:lem:alpha_r_prob1b_eq1},
it can be shown that for $ \frac{1}{\sqrt{N}} \leq r \leq 1$,
\[ b \leq \frac{1}{N}  \leq s_1 \leq a \leq 1. \]
Hence, the optimal solution to problem~\eqref{eq:lem:alpha_r_prob1} is simply $s_1^\star = a$,
and the proof is complete.

}

\bibliographystyle{IEEEtran}
\bibliography{ref_mves}

\begin{thebibliography}{10}
\providecommand{\url}[1]{#1}
\csname url@samestyle\endcsname
\providecommand{\newblock}{\relax}
\providecommand{\bibinfo}[2]{#2}
\providecommand{\BIBentrySTDinterwordspacing}{\spaceskip=0pt\relax}
\providecommand{\BIBentryALTinterwordstretchfactor}{4}
\providecommand{\BIBentryALTinterwordspacing}{\spaceskip=\fontdimen2\font plus
\BIBentryALTinterwordstretchfactor\fontdimen3\font minus
  \fontdimen4\font\relax}
\providecommand{\BIBforeignlanguage}[2]{{%
\expandafter\ifx\csname l@#1\endcsname\relax
\typeout{** WARNING: IEEEtran.bst: No hyphenation pattern has been}%
\typeout{** loaded for the language `#1'. Using the pattern for}%
\typeout{** the default language instead.}%
\else
\language=\csname l@#1\endcsname
\fi
#2}}
\providecommand{\BIBdecl}{\relax}
\BIBdecl

\bibitem{bioucas13overview}
J.~M. Bioucas-Dias, A.~Plaza, G.~Camps-Valls, P.~Scheunders, N.~Nasrabadi, and
  J.~Chanussot, ``Hyperspectral remote sensing data analysis and future
  challenges,'' \emph{IEEE Geosci. Remote Sens. Mag.}, vol.~1, no.~2, pp.
  6--36, Jun. 2013.

\bibitem{14SPM}
W.-K. Ma, J.~M. Bioucas-Dias, J.~Chanussot, and P.~Gader, Eds., \emph{Special
  Issue on Signal and Image Processing in Hyperspectral Remote Sensing, IEEE
  Signal Process. Mag.}, vol.~31, no.~1, Jan. 2014.

\bibitem{Jose12}
J.~Bioucas-Dias, A.~Plaza, N.~Dobigeon, M.~Parente, Q.~Du, P.~Gader, and
  J.~Chanussot, ``Hyperspectral unmixing overview: {G}eometrical, statistical,
  and sparse regression-based approaches,'' \emph{IEEE J. Sel. Topics Appl.
  Earth Observ.}, vol.~5, no.~2, pp. 354--379, 2012.

\bibitem{Ken14SPM_HU}
W.-K. Ma, J.~M. Bioucas-Dias, T.-H. Chan, N.~Gillis, P.~Gader, A.~J. Plaza,
  A.~Ambikapathi, and C.-Y. Chi, ``A signal processing perspective on
  hyperspectral unmixing,'' \emph{IEEE Signal Process. Mag.}, vol.~31, no.~1,
  pp. 67--81, 2014.

\bibitem{Dobigeon09}
N.~Dobigeon, S.~Moussaoui, M.~Coulon, J.-Y. Tourneret, and A.~O. Hero, ``Joint
  {B}ayesian endmember extraction and linear unmixing for hyperspectral
  imagery,'' \emph{IEEE Trans. Signal Process.}, vol.~57, no.~11, pp.
  4355--4368, Nov. 2009.

\bibitem{Chan2009}
T.-H. Chan, C.-Y. Chi, Y.-M. Huang, and W.-K. Ma, ``A convex analysis based
  minimum-volume enclosing simplex algorithm for hyperspectral unmixing,''
  \emph{IEEE Trans. Signal Process.}, vol.~57, no.~11, pp. 4418--4432, 2009.

\bibitem{gillis2014fast}
N.~Gillis and S.~A. Vavasis, ``Fast and robust recursive algorithms for
  separable nonnegative matrix factorization,'' \emph{IEEE Trans. Pattern Anal.
  Mach. Intell.}, vol.~36, no.~4, pp. 698--714, 2014.

\bibitem{Li2008}
J.~Li and J.~Bioucas-Dias, ``Minimum volume simplex analysis: {A} fast
  algorithm to unmix hyperspectral data,'' in \emph{Proc. {IEEE} IGARSS}, Aug.
  2008.

\bibitem{Craig1994}
M.~D. Craig, ``Minimum-volume transforms for remotely sensed data,''
  \emph{{IEEE} Trans. Geosci. Remote Sens.}, vol.~32, no.~3, pp. 542--552, May
  1994.

\bibitem{Full81}
W.~E. Full, R.~Ehrlich, and J.~E. Klovan, ``{EXTENDED QMODEL}---objective
  definition of external endmembers in the analysis of mixtures,''
  \emph{Mathematical Geology}, vol.~13, no.~4, pp. 331--344, 1981.

\bibitem{Winter1999}
M.~E. Winter, ``N-findr: An algorithm for fast autonomous spectral end-member
  determination in hyperspectral data,'' in \emph{Proc. {SPIE} Conf. Imaging
  Spectrometry}, Pasadena, {CA}, Oct. 1999, pp. 266--275.

\bibitem{du2008end}
Q.~Du, N.~Raksuntorn, N.~H. Younan, and R.~L. King, ``End-member extraction for
  hyperspectral image analysis,'' \emph{Applied Optics}, vol.~47, no.~28, pp.
  F77--F84, 2008.

\bibitem{Chan2011}
T.-H. Chan, W.-K. Ma, A.~Ambikapathi, and C.-Y. Chi, ``A simplex volume
  maximization framework for hyperspectral endmember extraction,'' \emph{IEEE
  Trans. Geosci. Remote Sens.}, vol.~49, no.~11, pp. 4177--4193, 2011.

\bibitem{miao2007endmember}
L.~Miao and H.~Qi, ``Endmember extraction from highly mixed data using minimum
  volume constrained nonnegative matrix factorization,'' \emph{IEEE Trans.
  Geosci. Remote Sens.}, vol.~45, no.~3, pp. 765--777, 2007.

\bibitem{agathos2014gpu}
A.~Agathos, J.~Li, D.~Petcu, and A.~Plaza, ``Multi-{GPU} implementation of the
  minimum volume simplex analysis algorithm for hyperspectral unmixing,''
  \emph{{\rm to appear in } IEEE J. Sel. Topics Appl. Earth Observ.}, 2014.

\bibitem{Dias2009}
J.~Bioucas-Dias, ``A variable splitting augmented {L}agrangian approach to
  linear spectral unmixing,'' in \emph{Proc. IEEE WHISPERS}, Aug. 2009.

\bibitem{Arul2011}
A.~Ambikapathi, T.-H. Chan, W.-K. Ma, and C.-Y. Chi, ``Chance-constrained
  robust minimum-volume enclosing simplex algorithm for hyperspectral
  unmixing,'' \emph{IEEE Trans. Geosci. Remote Sens.}, vol.~49, no.~11, pp.
  4194--4209, 2011.

\bibitem{hendrix2012new}
E.~M. Hendrix, I.~Garc{\'\i}a, J.~Plaza, G.~Martin, and A.~Plaza, ``A new
  minimum-volume enclosing algorithm for endmember identification and abundance
  estimation in hyperspectral data,'' \emph{IEEE Trans. Geosci. Remote Sens.},
  vol.~50, no.~7, pp. 2744--2757, 2012.

\bibitem{Lopes2010}
M.~B. Lopes, J.~C. Wolff, J.~Bioucas-Dias, and M.~Figueiredo, ``{NIR}
  hyperspectral unmixing based on a minimum volume criterion for fast and
  accurate chemical characterisation of counterfeit tablets,'' \emph{Analytical
  Chemistry}, vol.~82, no.~4, pp. 1462--1469, 2010.

\bibitem{nascimento2012hyperspectral}
J.~Nascimento and J.~Bioucas-Dias, ``Hyperspectral unmixing based on mixtures
  of {Dirichlet} components,'' \emph{IEEE Trans. Geosci. Remote Sens.},
  vol.~50, no.~3, pp. 863--878, 2012.

\bibitem{plaza2012endmember}
J.~Plaza, E.~M. Hendrix, I.~Garc{\'\i}a, G.~Mart{\'\i}n, and A.~Plaza, ``On
  endmember identification in hyperspectral images without pure pixels: A
  comparison of algorithms,'' \emph{Journal of Mathematical Imaging and
  Vision}, vol.~42, no. 2-3, pp. 163--175, 2012.

\bibitem{lin2013end}
C.-H. Lin, A.~Ambikapathi, W.-C. Li, and C.-Y. Chi, ``On the endmember
  identifiability of {Craig's} criterion for hyperspectral unmixing: {A}
  statistical analysis for three-source case,'' in \emph{Proc. IEEE ICASSP},
  May 2013, pp. 2139--2143.

\bibitem{boyd2004convex}
S.~Boyd and L.~Vandenberghe, \emph{Convex Optimization}.\hskip 1em plus 0.5em
  minus 0.4em\relax Cambridge University Press, 2004.

\bibitem{gritzmann1995largestj}
P.~Gritzmann, V.~Klee, and D.~Larman, ``Largest $j$-simplices in
  $n$-polytopes,'' \emph{Discrete and Computational Geometry}, vol.~13, no.~1,
  pp. 477--515, 1995.

\bibitem{packer2002np}
A.~Packer, ``{NP}-hardness of largest contained and smallest containing
  simplices for {V}- and {H}-polytopes,'' \emph{Discrete and Computational
  Geometry}, vol.~28, no.~3, pp. 349--377, 2002.

\bibitem{gritzmann1994complexity}
P.~Gritzmann and V.~Klee, ``On the complexity of some basic problems in
  computational convexity: {I}. containment problems,'' \emph{Discrete
  Mathematics}, vol. 136, no.~1, pp. 129--174, 1994.

\bibitem{Chan2008}
T.-H. Chan, W.-K. Ma, C.-Y. Chi, and Y.~Wang, ``A convex analysis framework for
  blind separation of non-negative sources,'' \emph{IEEE Trans. Signal
  Process.}, vol.~56, no.~10, pp. 5120--5134, 2008.

\bibitem{USGS2007}
R.~Clark, G.~Swayze, R.~Wise, E.~Livo, T.~Hoefen, R.~Kokaly, and S.~Sutley,
  ``{USGS} digital spectral library splib06a: {U.S. Geological Survey, Digital
  Data Series 231},'' \url{http://speclab.cr.usgs.gov/spectral.lib06}, 2007.

\bibitem{gerber1975}
L.~Gerber, ``The orthocentric simplex as an extreme simplex,'' \emph{Pacific
  Journal of Mathmatics}, vol.~56, no.~1, pp. 97--111, Nov. 1975.

\end{thebibliography}

\end{document}